\documentclass[10pt,journal,compsoc]{IEEEtran}
\usepackage{cite}
\usepackage{amsmath,amssymb,amsfonts}
\usepackage{algorithm,algorithmic}
\usepackage{graphicx}
\usepackage{textcomp}
\usepackage{comment}
\usepackage{threeparttable}
\usepackage{hyperref}
\usepackage{threeparttable}
\usepackage{multirow}
\usepackage{booktabs}
\usepackage{subfig}

\def\BibTeX{{\rm B\kern-.05em{\sc i\kern-.025em b}\kern-.08em
    T\kern-.1667em\lower.7ex\hbox{E}\kern-.125emX}}

\begin{document}
\title{A Masked Semi-Supervised Learning Approach for Otago Micro Labels Recognition}
\author{Meng~Shang,
        Lenore~Dedeyne,
        Jolan~Dupont,
		Laura~Vercauteren,
		Nadjia~Amini,
		Laurence~Lapauw,
		Evelien~Gielen,
		Sabine~Verschueren,
		Carolina~Varon,
		Walter~De Raedt,
        and~Bart~Vanrumste% <-this % stops a space
\thanks{M. Shang, C. Varon and B. Vanrumste are with KU Leuven, STADIUS, Department of Electrical Engineering, 3000 Leuven, Belgium, e-mail: meng.shang@kuleuven.be.}% <-this % stops a space
\thanks{M. Shang and W. De Raedt are with Imec, Kapeldreef 75, 3001 Leuven, Belgium.}% <-this % stops a space
\thanks{M. Shang and B. Vanrumste are with KU Leuven, e-Media Research lab.}
\thanks{L. Dedeyne, J. Dupont, L. Vercauteren, N. Amini, L. Lapauw, E. Gielen are with Geriatrics \& Gerontology, Department of Public Health and Primary Care, KU Leuven, Belgium.}
\thanks{J. Dupont and E. Gielen are with Department of Geriatric Medicine, UZ Leuven, Belgium.}
\thanks{S. Verschueren is with the Musculoskeletal Rehabilitation Research Group, Department of Rehabilitation Sciences, KU Leuven.}
}

\maketitle

\begin{abstract}
The Otago Exercise Program (OEP) serves as a vital rehabilitation initiative for older adults, aiming to enhance their strength and balance, and consequently prevent falls. While Human Activity Recognition (HAR) systems have been widely employed in recognizing the activities of individuals, existing systems focus on the duration of macro activities (i.e. a sequence of repetitions of the same exercise), neglecting the ability to discern micro activities (i.e. the individual repetitions of the exercises), in the case of OEP. This study presents a novel semi-supervised machine learning approach aimed at bridging this gap in recognizing the micro activities of OEP. To manage the limited dataset size, our model utilizes a Transformer encoder for feature extraction, subsequently classified by a Temporal Convolutional Network (TCN). Simultaneously, the Transformer encoder is employed for masked unsupervised learning to reconstruct input signals. Results indicate that the masked unsupervised learning task enhances the performance of the supervised learning (classification task), as evidenced by f1-scores surpassing the clinically applicable threshold of 0.8. From the micro activities, two clinically relevant outcomes emerge: counting the number of repetitions of each exercise and calculating the velocity during \textit{chair rising}. These outcomes enable the automatic monitoring of exercise intensity and difficulty in the daily lives of older adults.
\end{abstract}


\section{Introduction}
\label{sec:introduction}
\IEEEPARstart {O}tago Exercise Program (OEP) is a rehabilitation initiative designed for older adults with the aim of enhancing their strength and balance, thereby reducing the risk of falls \cite{thomas_does_2010}. Consisting of over 17 strength and balance exercises and a walking plan, the program targets various functions of the body \cite{mat_effect_2018,almarzouki_improved_2020}. Typically, older adults within the community are encouraged to engage in OEP sessions two or three times per week. Subsequently, their adherence to the program is reported to therapists for in-depth analysis. In an effort to overcome the limitations associated with self-reports, several studies have explored the application of wearable sensors to monitor the execution of OEP \cite{shang2023otago, dedeyne_exploring_2021, bevilacqua_human_2019, shang2024ds}. \par

In a preceding investigation \cite{shang2023otago}, we introduced a hierarchical system utilizing a single wearable Inertial Measurement Unit (IMU) to monitor the Otago Exercise Program (OEP) in the daily lives of older adults. This system demonstrated the ability to identify the start and end of the program within the context of activities of daily living (ADLs), achieving f1-scores larger than 0.95. Additionally, specific exercise types were recognizable with f1-scores surpassing 0.8. Subsequently, in another study \cite{shang2024ds}, we proposed a more robust sequence-to-sequence model aimed at refining the recognition of distinct OEP exercise types. Throughout these research endeavors, our analyses yielded the recognition of four macro activities inherent in the OEP within participants' daily routines: \textit{ankle plantarflexors}, \textit{knee bends}, \textit{abdominal muscles exercise}, and \textit{chair rising}. A macro activity was defined as a series of repetitive movements constituting the same exercise (e.g. five repetitions of chair rising). \par

Although the therapists could specify the duration subjects spend on a particular type of OEP exercises to assess their compliance, there are still two questions that need to be explored: 

\begin{itemize}
	\item What is the count of repetitions for each exercise they performed? This provides supplementary insights into their workload. As subjects enhance their fitness levels, there should be a corresponding increase in the number of exercise repetitions \cite{thomas_does_2010, kocic2018effectiveness}. Hence, it is crucial to track the repetition count, as it is indicative of the exercises' difficulty.
	\item Is it feasible to automatically derive clinical parameters from each repetition of \textit{chair rising}? It is the most common activity not only in OEP but also in daily life, including \textit{sit-to-stand} and \textit{stand-to-sit}. From a clinical perspective, scrutinizing the quality of each \textit{chair rising} repetition yields more meaningful outcomes than focusing solely on its duration \cite{martinez2019probabilistic}. The quality, encompassing factors like acceleration and velocity, holds valuable insights \cite{adamowicz2020assessment, marques2020accelerometer}. Once the acceleration signals for each \textit{chair rising} repetition are extracted, these clinical parameters can be computed.
	
\end{itemize}

To address these two questions, it is essential to recognize each repetition of the OEP exercises. In our previous study \cite{shang2024ds}, the repetitions were defined as micro activities. Despite the enhancement in recognizing macro activities facilitated by micro activities, the stand-alone recognition performance of micro activities remained unsatisfactory. Therefore, this study aims to develop a system based on a wearable IMU to recognize the micro activities of OEP. \par

One of the challenges of recognizing micro activities is the limited size of the annotated dataset \cite{shang2024ds}. Annotating each exercise repetition as a segment necessitates a substantial workload, especially for deep learning models that are greedy for data \cite{soekhoe2016impact}. To enhance results in the face of limited training samples, a solution was devised based on a semi-supervised architecture \cite{hermans2023multi}. The architecture encompasses two parts of the learning process: the first part involves the classification (supervised) task with the labeled data, while the second part complies with the regression (unsupervised) task using the unlabeled data. The regression task was designed to help the model better understand the input context and improve the classification performance.\par

This study proposes a system based on a single wearable IMU to recognize the micro activities of OEP exercises. To make use of the limited labeled training data, a semi-supervised deep learning model was utilized, with an extra signal reconstruction block to improve the classification results. With the recognized micro activities, the number of repetitions can be counted and the velocity of \textit{chair rising} can be automatically calculated as clinical outcomes.

The contributions of this study are:
\begin{itemize}
	\item This study utilizes a single wearable sensor to recognize the repetitions (micro activities) of OEP exercises for older adults in their daily lives. Five types of micro activities can be recognized with f1-scores higher than 0.8. To our knowledge, this is the first study recognizing each repetition of a rehabilitation program using wearable sensors.
	\item With the recognized micro activities, two clinical outcomes can be extracted. Firstly, the counting of repetitions for each exercise provides valuable insights into subjects' compliance with the program. Secondly, the calculation of the velocity during \textit{chair rising} serves as an indicator of the activity's intensity.
	\item This study applies a semi-supervised approach, with an extra reconstruction block to improve the classification results for OEP recognition. The approach showed outperforming performance with the limited dataset for training. The method also validated the usage of masked autoencoders in HAR systems.
\end{itemize}

The structure of this paper is organized as follows: Section~\ref{sec:related works} provides a review of the current state of HAR systems. Section~\ref{sec:method} details the datasets, elucidates the implementation process, and validates the proposed system. Section~\ref{sec:results} presents the experimental findings, while Section~\ref{sec:discussion} delves into a thorough discussion of these results. Finally, Section~\ref{sec:conclusion} concludes the paper and suggests potential directions for future research.

\section{Related works}
\label{sec:related works}

\subsection{Deep Learning in HAR}
\label{subsec:related works DL}

Traditionally, HAR systems relied on machine learning models that employed hand-crafted features \cite{wang_survey_2019}. However, in recent years, deep learning has emerged as a more efficient alternative for time series classification in HAR systems. Among the network-based models, Convolutional Neural Networks (CNNs) are utilized for extracting neighboring information from adjacent input \cite{8684824, lee_human_2017, wagner_activity_2017}, while Recurrent Neural Networks (RNNs) are employed for capturing temporal patterns \cite{zhao_deep_2018}. Furthermore, more sophisticated models have been developed by combining these basic structures. For instance, the CNN-LSTM model integrates CNNs with Long-Short Term Memory (LSTM) structures to leverage diverse input characteristics \cite{mutegeki_cnn-lstm_2020, mekruksavanich_smartwatch-based_2020}. Additionally, Temporal Convolutional Networks (TCNs) have been constructed as an extension of CNNs \cite{farha_ms-tcn_2019}. Their advantage lies in the ability to extract both long-term and short-term features by extending dilation factors without encountering issues like gradient explosion.\par

Besides these explored models, Transformers have garnered attention in many domains. Originally developed for natural language processing (NLP) \cite{vaswani2017attention, vig2019analyzing, wang2019language}, Transformers utilize an attention mechanism to calculate scores (similarity) between the query vector of a target value and all key vectors of the input sequence. This mechanism ensures that, during output generation, the model is cognizant of which samples from the input require more focus. While initially designed for NLP, Transformers have found applications in image processing \cite{chen2021pre}, speech processing \cite{dong2018speech}, and time series processing \cite{10409509}. Notably, there have been several recent studies applying this architecture to HAR \cite{dirgova2022wearable}.\par

\subsection{Semi-supervised learning}

HAR systems conventionally fall under the category of supervised learning, primarily designed for classification tasks. Nevertheless, the exploration of unsupervised learning approaches has also been pursued to enhance classification performance. In unsupervised learning for HAR, the common task involves reconstructing input signals using autoencoders \cite{mohd2021feature, thill2021temporal}. The latent features derived from this reconstruction process can subsequently be employed for classification purposes. \par

The most popular unsupervised architecture is CNN autoencoders \cite{mohd2021feature, seyfiouglu2018deep}. In this design, convolutional layers are responsible for feature compression, while subsequent deconvolution layers handle the reconstruction of input signals. Similarly, LSTM autoencoders capture learned representations through an LSTM encoder layer and then expand this representation to generate the input sequence via another LSTM decoder layer \cite{nguyen2021forecasting}. Furthermore, derivatives such as TCN autoencoders \cite{thill2021temporal}, CNN-LSTM autoencoders \cite{8422895}, among others, have also been explored.\par

\begin{figure}[!h]
\centering
\includegraphics[width=\columnwidth]{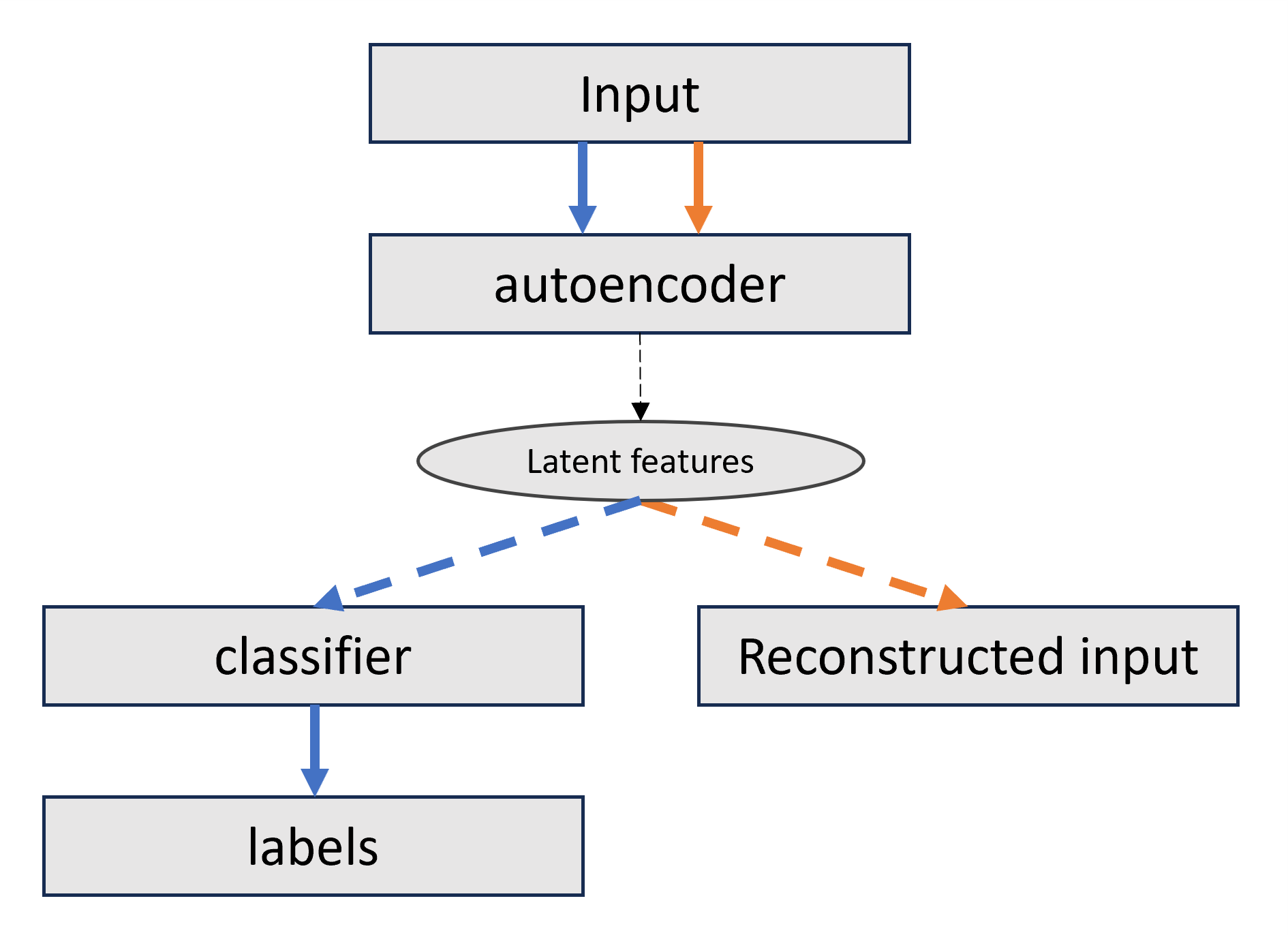}
\caption{The architecture of a typical semi-supervised method. The autoencoder reconstructs the input while the latent features were also applied for classification. The loss function is the combination of both routes.}
\label{fig:semi structure}
\end{figure}

A type of semi-supervised learning architecture was proposed incorporating both supervised and unsupervised tasks \cite{hermans2023multi, yu2023semi}. This approach trains the model through two simultaneous processes: reconstructing the input and classifying the labels. The reconstruction part is only employed to enhance the model's comprehension of the input, thereby improving classification performance. An advantageous feature of this approach is its ability to accommodate multiple tasks in a single step, combining different loss functions. The architecture of a typical semi-supervised method is shown in Fig.~\ref{fig:semi structure}. 

\subsection{Masked autoencoders}

Through the process of reconstructing the input, autoencoders acquire features crucial for classification. However, there is a potential limitation where the reconstruction tasks might be overly simplistic, resulting in less informative features. In the realm of NLP, the BERT model \cite{nozza2020mask, wettig2022should} has gained popularity for addressing this challenge through masked reconstruction. This strategy involves masking or replacing certain input words, enabling the model to glean richer contextual information. \par

In 2022, He et al. extended masked autoencoders for image reconstruction \cite{he2022masked}. In this extension, during the reconstruction of images, certain patches were masked in the input and subsequently reconstructed in the output. They advocated for a high mask ratio (75\%) given the less dense information inherent in images compared to natural language. This approach demonstrated superior classification performance when compared to supervised approaches and other unsupervised models.\par

Following these studies, the same strategy has been extensively adopted for time series reconstruction across various domains, encompassing EEG \cite{chien2022maeeg} and ECG \cite{10409509}. Notably, in HAR systems, the masked transformer has demonstrated superior results compared to both supervised methods and other autoencoders, even without pre-training. \cite{haresamudram2020masked}.

\section{Materials and Methodology}
\label{sec:method}

\subsection{Dataset Collection}

This study received approval from the Ethics Committee Research UZ/KU Leuven (S59660 and S60763). Written informed consent was obtained from all participants prior to study participation. This follows up on our previous studies \cite{dedeyne_exploring_2021, shang2023otago, shang2024ds} that explored the same dataset. \par

Within the experimental framework, a cohort of community-dwelling older adults aged 65 years and above, engaged in a modified rendition of the Otago Exercise Program (OEP) while utilizing the McRoberts MoveMonitor (McRoberts B.V., Netherlands). This monitoring device was equipped with an IMU positioned at the waist. The OEP exercises comprise multiple sub-classes, details of which are outlined in the previous study \cite{shang2023otago}. In addition to OEP sessions, the subjects also undertook various ADLs. There were two scenarios for dataset collection with different subjects:

\subsubsection{Lab scenario}
The dataset was collected in the lab from 35 subjects. These subjects performed OEP and ADLs under the guidance of researchers certified as OEP leaders. The ADLs included walking, stair climbing, sitting, standing, and indoor cycling.

\subsubsection{Home scenario}
This dataset was collected at home with video recordings of seven subjects. With the camera on, the subjects followed a booklet containing instructions to perform OEP. They also performed ADLs before and/or after OEP without any instructions or camera recordings.

For both scenarios, the recruited older adults were classified as either (pre-)sarcopenic or non-sarcopenic, as defined by EWGSOP1 \cite{cruz2010sarcopenia}. Detailed information is provided in Table~\ref{tab:subinfo}. It is important to note that the subjects recruited for the two datasets differed. During intervals between exercises, participants were given autonomy: no specific instructions or monitoring were implemented.  \par

\begin{table}[!h]
\caption{Information of the subjects}
\centering
\label{tab:subinfo}
\begin{tabular}{|l|l|l|ll|}
\hline
     & number & age                                                       & \multicolumn{2}{l|}{gender \&   sarcopenia}                                                                                      \\ \hline
Lab  & 36     & \begin{tabular}[c]{@{}l@{}}79.33$\pm$\\ 5.73\end{tabular} & \multicolumn{2}{l|}{\begin{tabular}[c]{@{}l@{}}17 females (7 (pre-)sarcopenia),\\ 19   males (11 (pre-)sarcopenia)\end{tabular}} \\ \hline
Home & 7      & \begin{tabular}[c]{@{}l@{}}69.43$\pm$\\ 2.92\end{tabular} & \multicolumn{2}{l|}{\begin{tabular}[c]{@{}l@{}}4 females (2 (pre-)sarcopenia),\\ 3 males (2 (pre-)sarcopenia)\end{tabular}}      \\ \hline
\end{tabular}
\end{table}

According to the previous study \cite{shang2023otago, shang2024ds}, only limited sub-classes were recognizable using a waist-mounted IMU. Therefore, four Otago exercises were explored for classification in this study: \textit{ankle plantarflexors}, \textit{knee bend}, \textit{abdominal muscles exercise}, and \textit{chair rising}.

\subsection{micro labels annotation}

As defined in \cite{shang2024ds}, the micro labels are the repetitions of the OEP exercises. Different from the traditional macro labels, the intervals between repetitions were not labeled as micro labels. In this study, to better differentiate the micro and macro activities, the names and definitions of micro activities are as follows:

\begin{itemize}
	\item \textit{heels up/down} (the micro activity of \textit{ankle plantarflexors}): start when the heels leave the ground and end when they touch the ground.
	\item \textit{knees flexion/extension} (the micro activity of \textit{knee bends}): start when the knees bend and end when they become straight
	\item \textit{Trunk flexion/extension} (the micro activity of \textit{abdominal muscles exercise}): start when the trunk leans backward and end when it becomes straight
	\item \textit{sit-to-stand} (the micro activity of \textit{chair rising}): start when the hip leaves the chair and end when standing straight.
	\item \textit{stand-to-sit} (the micro activity of \textit{chair rising}): start when the knees bend and end when taking the seat.
\end{itemize}

\begin{figure*}[!h]
\centering
\subfloat[heels up/down]{\includegraphics[width=0.9\columnwidth]{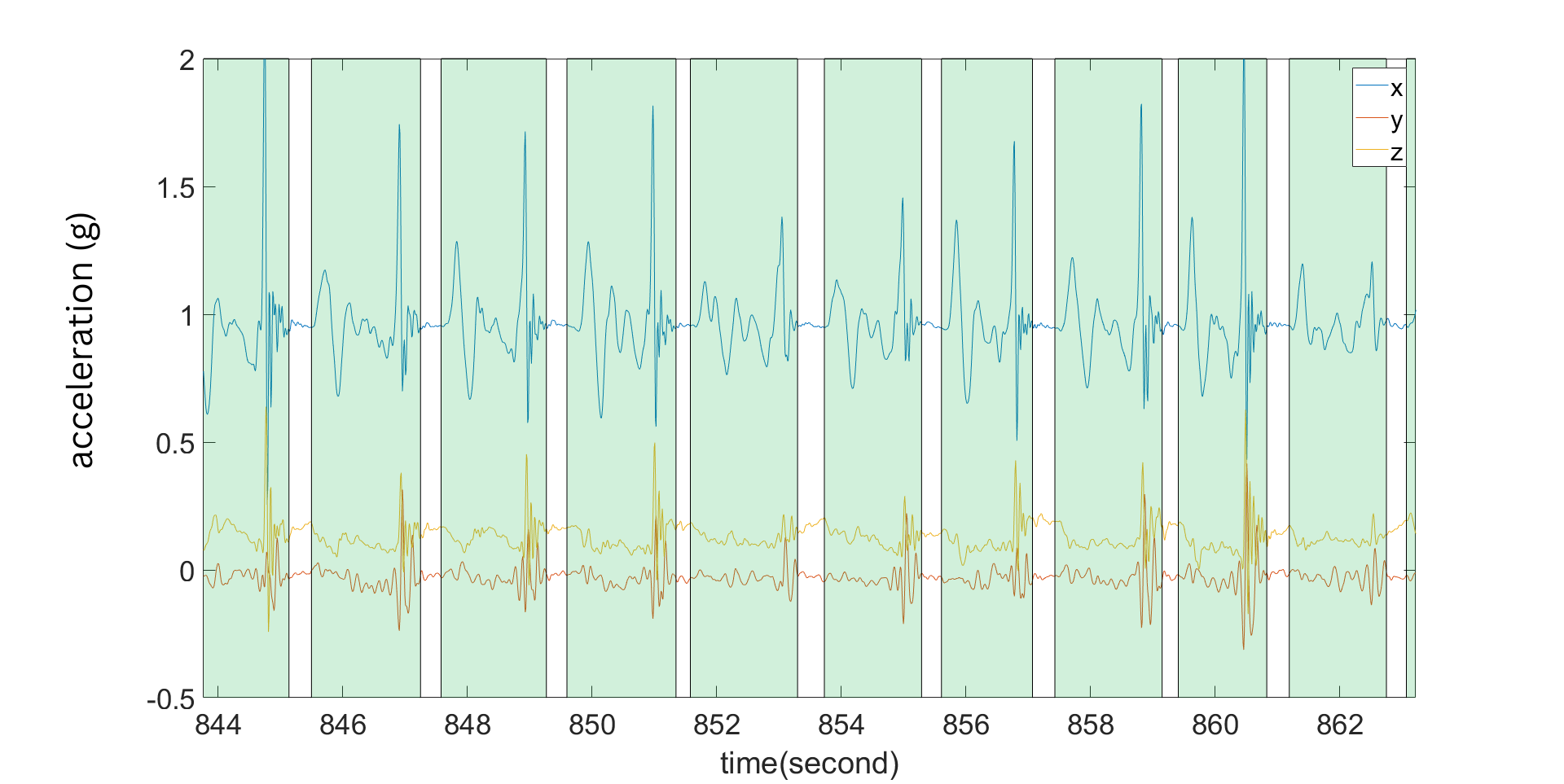}%
}
\subfloat[trunk flexion/extension]{\includegraphics[width=0.9\columnwidth]{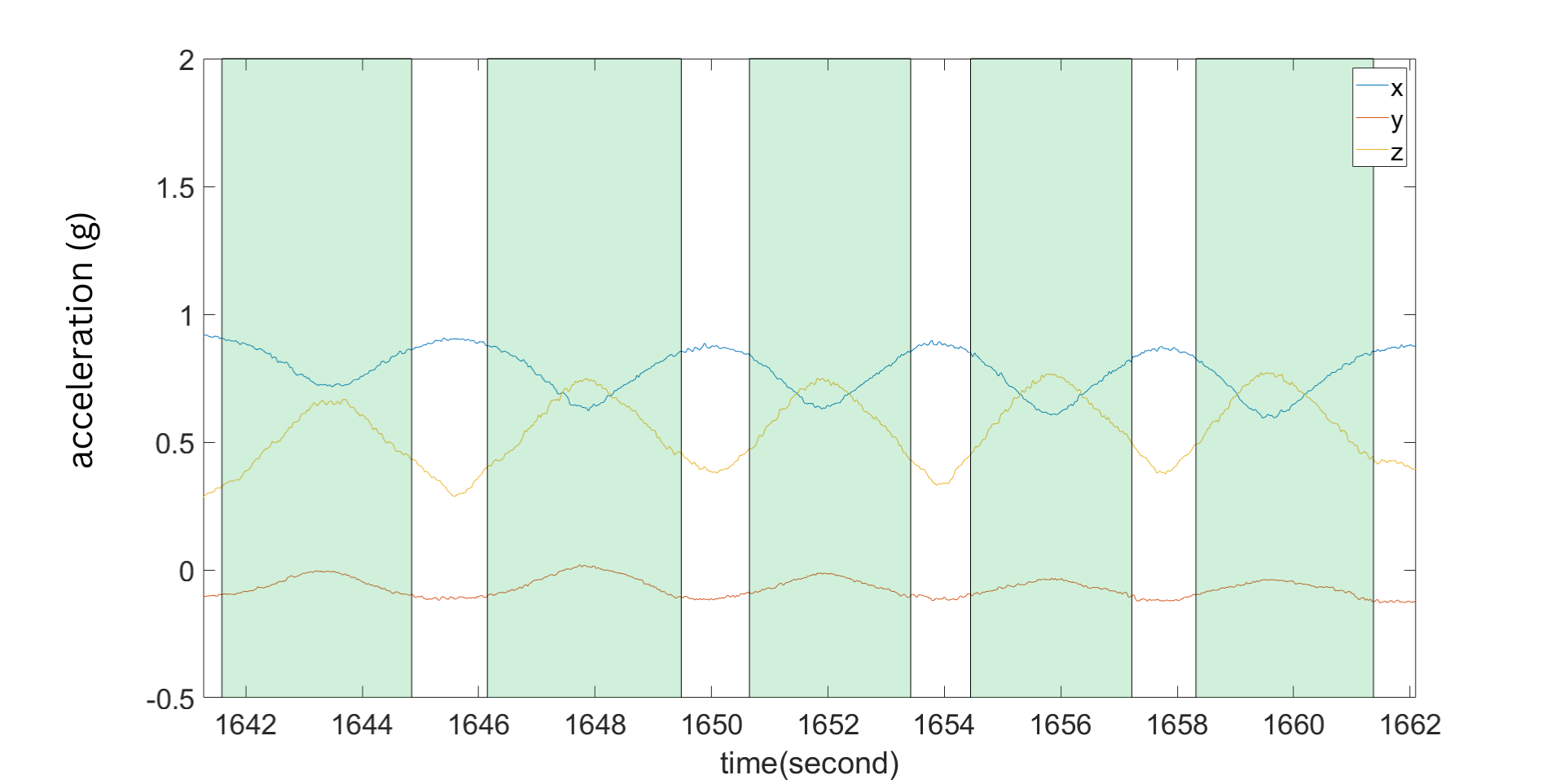}%
}
\hfill
\subfloat[knees flexion/extension]{\includegraphics[width=0.9\columnwidth]{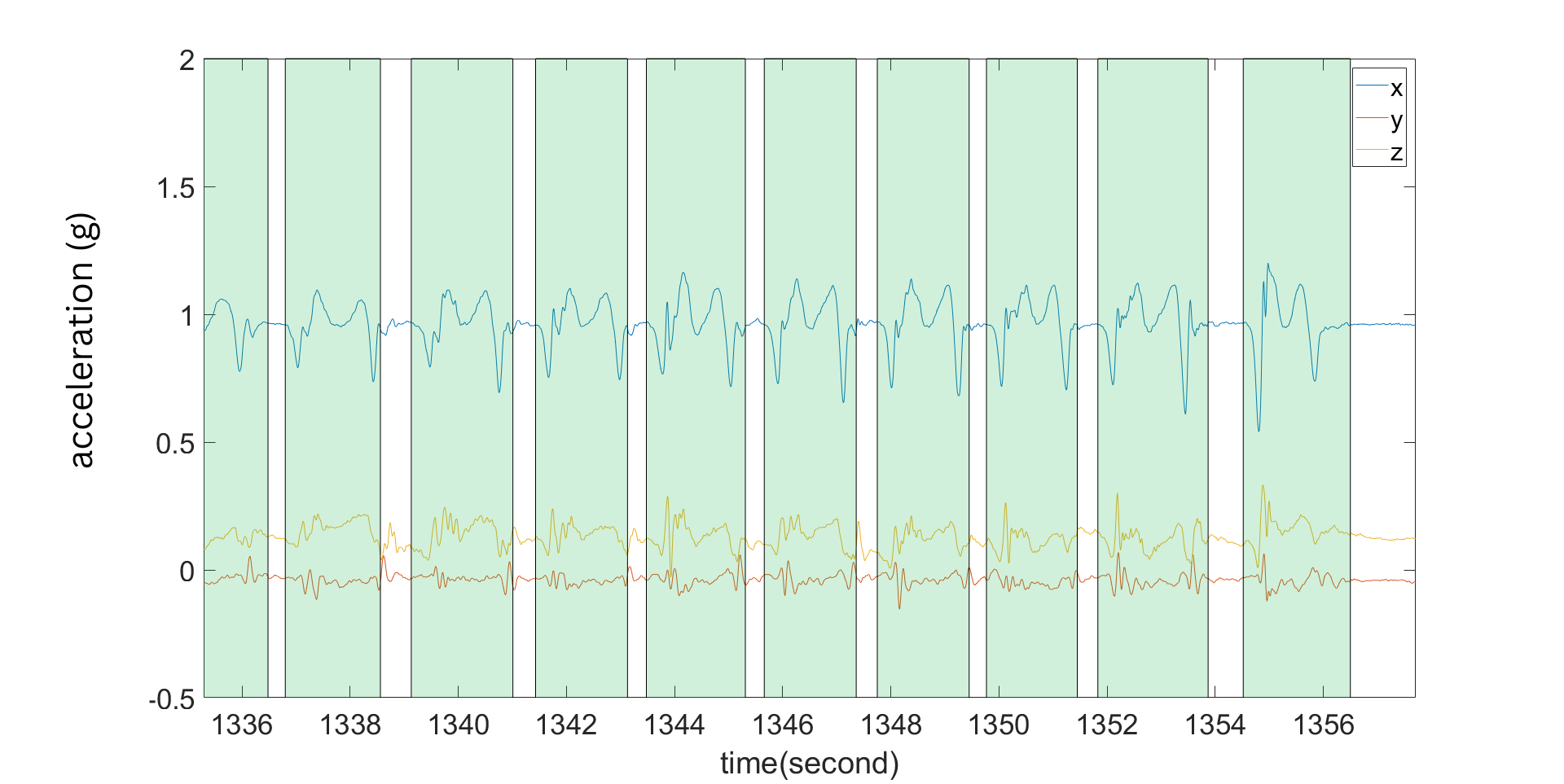}%
}
\subfloat[\textit{sit-to-stand} (green blocks) and \textit{stand-to-sit} (red blocks)]{\includegraphics[ width=0.9\columnwidth]{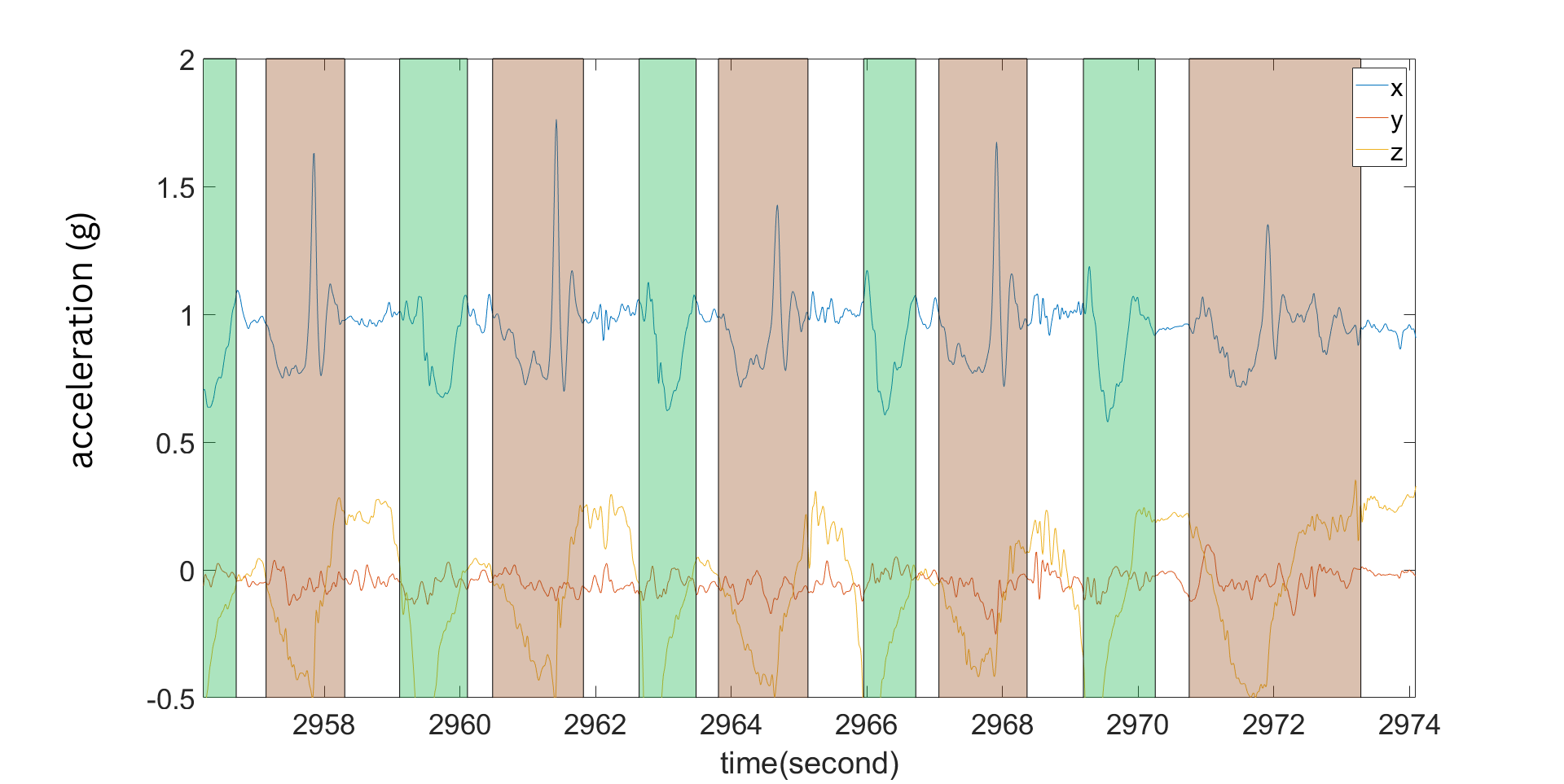}%
}
\caption{The annotated acceleration signals. The colored area illustrates the duration of a single repetition of OEP exercises.}
\label{fig:microlabels}
\end{figure*}

In this study, \textit{chair rising} was split into two types of micro activities: \textit{sit-to-stand} and \textit{stand-to-sit}. The reason was that the intervals of these two micro activities were large enough to manually annotate. The other micro activities could not be further split up based on the videos. Fig.~\ref{fig:microlabels} shows the annotated IMU signals of the micro activities. The numbers of labeled micro and macro segments are shown in Table~\ref{tab:number}.

\begin{table}[!h]
\caption{The number of micro and macro segments recorded}
\label{tab:number}
\center
\begin{tabular}{p{1.8cm}lll}
\hline
              & \begin{tabular}[c]{@{}l@{}}number of \\ micro segments\end{tabular} & \begin{tabular}[c]{@{}l@{}}number of\\ macro segments\end{tabular} & \begin{tabular}[c]{@{}l@{}}total duration \\ (min)\end{tabular} \\ \hline
heels up/down    & 970                                                                & 75                                                                   & 53.09                                                           \\ \hline
trunk flexion/extension     & 202                                                                & 35                                                                   & 16.94                                                           \\ \hline
knees flexion/extension    & 748                                                                & 75                                                                   & 33.92                                                           \\ \hline
sit-to-stand* & 369                                                                & 72                                                                   & 24.11                                                           \\ \hline
stand-to-sit* & 369                                                                & 72                                                                   & 24.11                                                           \\ \hline
\multicolumn{4}{l}{*\textit{chair rising} was split into \textit{sit-to-stand} and \textit{stand-to-sit}}                                                                                                                  
\end{tabular}
\end{table}

\subsection{Architecture overview}

\begin{figure}[!h]
\centering
\includegraphics[width=\columnwidth]{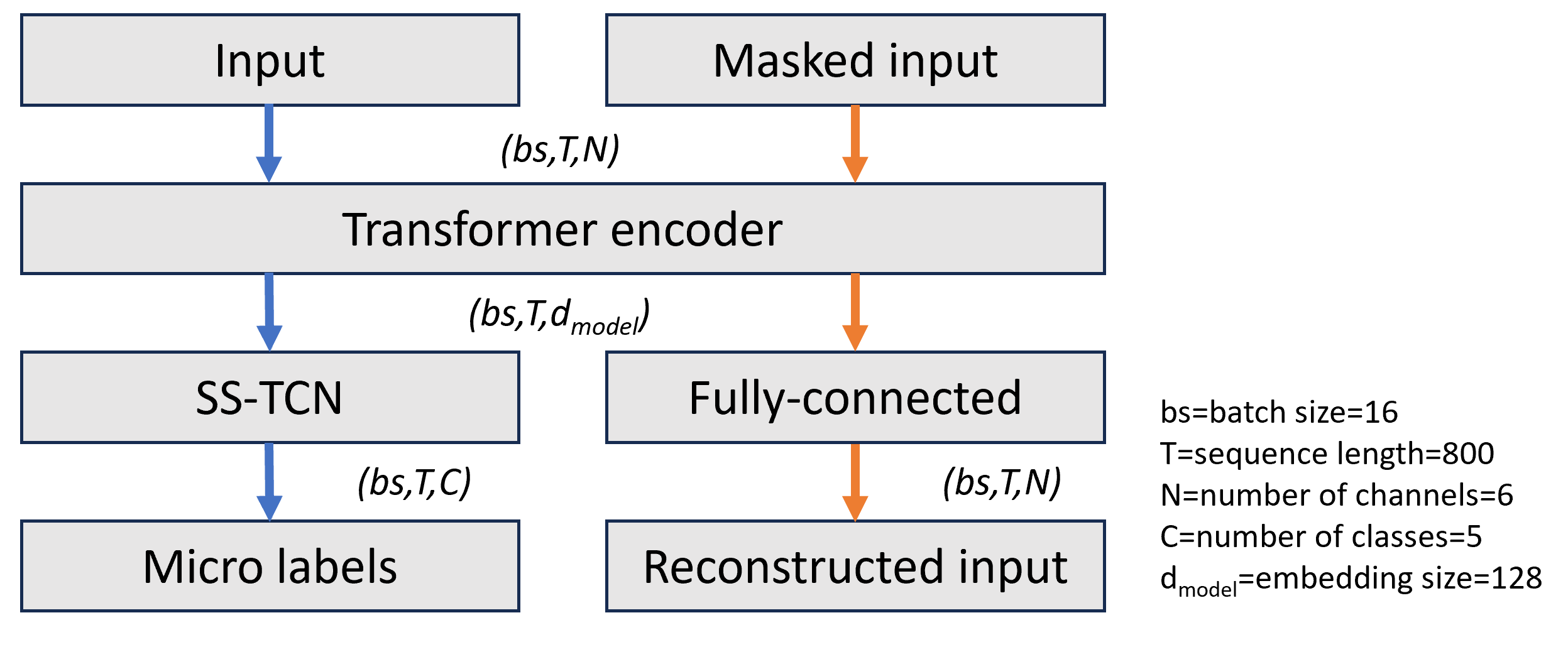}
\caption{The architecture of the proposed semi-supervised masked Transformer.}
\label{fig:transformer}
\end{figure}

In this study, inspired by the previous studies \cite{ibanez2022masked, haresamudram2020masked, hermans2023multi}, a semi-supervised masked transformer (Transformer SS-M) was introduced, as illustrated in Fig.~\ref{fig:transformer}. At the core of the model was the Transformer block, which engaged in two distinct tasks: classification and signal reconstruction. For classification, in the initial phase, the transformers extracted features from the raw input signals. Subsequently, a One-Stage Temporal Convolutional Network (OS-TCN) was deployed to classify these features into class predictions. Concurrently, the masked input signals were fed into the Transformer, and the extracted features were utilized for reconstructing the input signals through additional Fully-Connected Layers, as applied in the previous study \cite{haresamudram2020masked}. Importantly, the weights (parameters) of the Transformers were shared in both parts, ensuring that the output features were learned from both the supervised and unsupervised learning processes. \par

\subsection{Transformer}
\label{subsec:transformer}

\begin{figure}[!h]
\centering
\includegraphics[width=0.8\columnwidth]{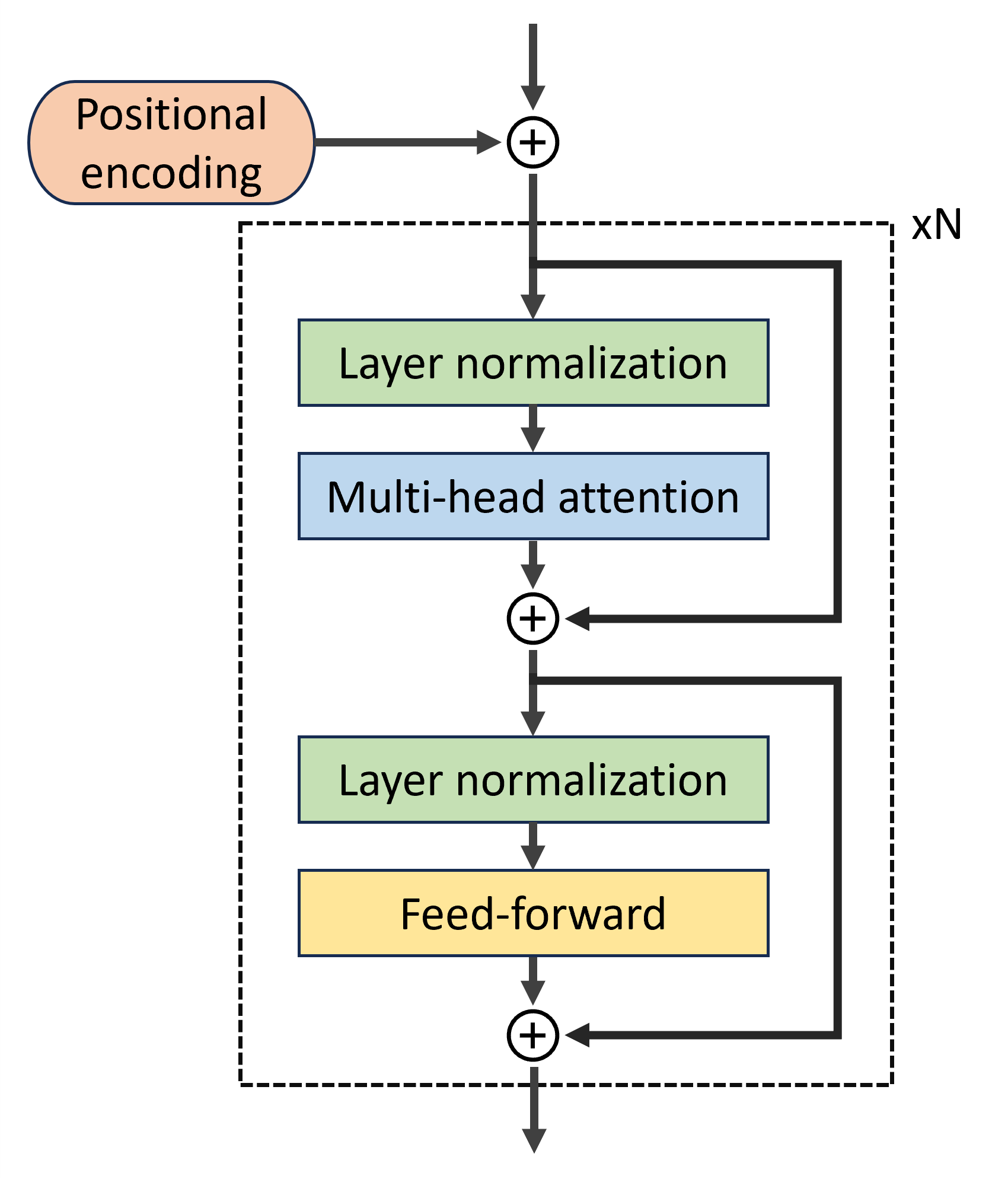}
\caption{The Transformer with multiple encoder blocks. }
\label{fig:encoder}
\end{figure}

The Transformer architecture has found widespread applications across diverse domains \cite{vaswani2017attention, vig2019analyzing, dirgova2022wearable}, as shown in Fig.~\ref{fig:encoder}. In this study, the input signals to the model incorporated only an accelerometer and a gyroscope. The magnetometer signals were excluded from utilization, as detailed in \cite{shang2023otago}. Consequently, the input to the model constituted a 6-axis time series.\par

Although the Transformer is capable of handling input sequences of variable lengths, practical considerations of computational cost led to the segmentation of the input using a fixed window with a length of 800 samples (8 seconds). Consequently, the input size was formatted as 16$\times$800$\times$6, where 16 represented the batch size.\par

The initial step involved mapping the input into a higher dimension through an embedding layer. Given that the model lacks inherent awareness of the sequence in the input time series, positional encoding was applied to incorporate sequential information. \par

A single Transformer encoder includes several types of layers. Following layer normalization, the multi-head attention layer was employed to extract similarities between input samples, with the self-attention mechanism being pivotal to the architecture \cite{vaswani2017attention}. This mechanism, facilitated by learnable parameters, computed three matrices: queries ($Q$) representing the information the model sought to attend, keys ($K$) representing information for comparison with queries, and values ($V$) used in the computation of the final attention function. The function is expressed as:

\begin{equation}
Attention = softmax(\frac{QK^T}{\sqrt{d_k}})V,
\end{equation}

where $d_k$ denotes the dimension of the keys. A multi-head attention layer comprised several heads, each equipped with individual queries, keys, and values to address diverse contexts. Subsequently, it was connected to feed-forward layers utilizing the rectified linear unit (RELU) function. The encoders were repeated and stacked to capture complex features.\par

The necessary hyperparameters of the applied Transformer in this study is shown in Table~\ref{tab:hyper}. The value of $d_k$ was hence $d_{model}/n_h = 16$, according to the definition in \cite{vaswani2017attention}.

\begin{table}[!h]
\caption{The hyperparameters of the applied Transformer}
\label{tab:hyper}
\centering
\begin{tabular}{ll}
\hline
embedding size ($d_{model}$)  & 128 \\ \hline
number of heads ($n_h$) & 8   \\ \hline
number of layers ($n_l$) & 3   \\ \hline
dropout rate     & 0.1 \\ \hline
\end{tabular}
\end{table}

It is noteworthy that Transformer decoders, a common component in most language models, were not utilized in this study \cite{vaswani2017attention}. As elucidated in \cite{zerveas2021transformer}, decoders are more suited for generative tasks with undefined output lengths. In the context of this study, the features generated by the encoders proved adequate for both time series classification and reconstruction, aligning with the approach in \cite{yu2023semi}. The details of reconstruction will be explained in section\ref{sec:unsupervised}. \par

\subsection{Supervised block}

\begin{figure}[!h]
\centering
\includegraphics[width=0.8\columnwidth]{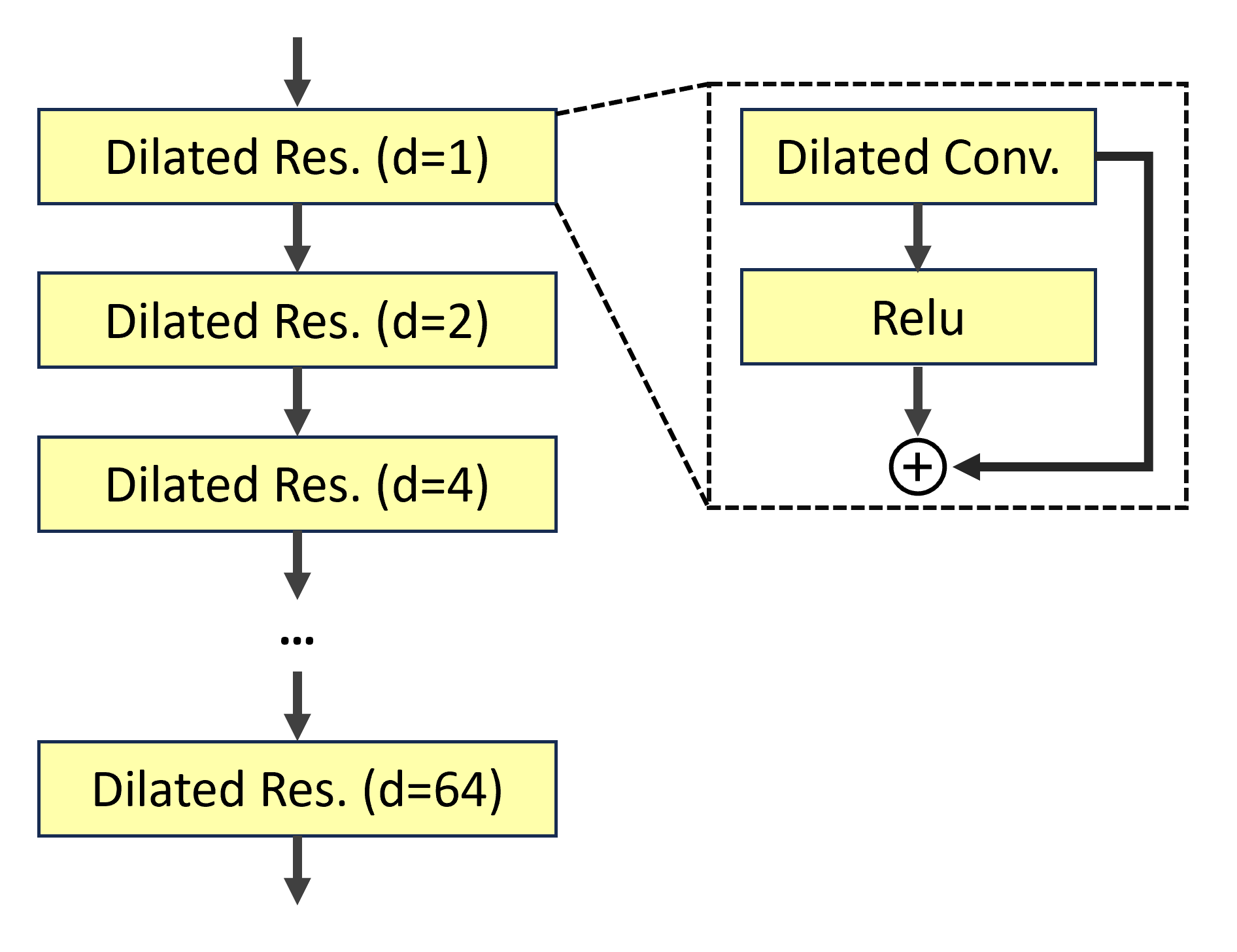}
\caption{The OS-TCN as the classifier.}
\label{fig:sstcn}
\end{figure}

The OS-TCN was employed to classify micro labels using features extracted by the Transformer. This classifier had been previously applied in our earlier study \cite{shang2024ds} with detailed implementations provided, as shown in Fig.~\ref{fig:sstcn}. The model consisted of stacked dilated convolutional layers, wherein the dilation factor increased exponentially with each forward layer—1, 2, 4, 8, and so forth. The use of stacked dilated layers offered two advantages. Firstly, it allowed the model to generate smoother predictions by producing an output with a long receptive field. Secondly, the training process remained unaffected by gradient explosion \cite{farha_ms-tcn_2019}. Since the micro activities were short, a receptive field of 255 samples (2.55 seconds) was achieved with seven stacked layers. The output of the OS-TCN yielded the classified micro labels.

The supervised loss function employed was the cross-entropy (CE) function:

\begin{equation}
L_{CE}=\frac{1}{TC}\sum_{t,c}-{y}_{t,c}\log{\hat{y}_{t,c}},
\end{equation}

where C denotes the total number of classes, ${y}_{t,c}$ and $\hat{y}_{t,c}$ denote the true and predicted label at time stamp $t$.

\subsection{Unsupervised block}
\label{sec:unsupervised}

\begin{figure}[!h]
\centering
\includegraphics[width=\columnwidth]{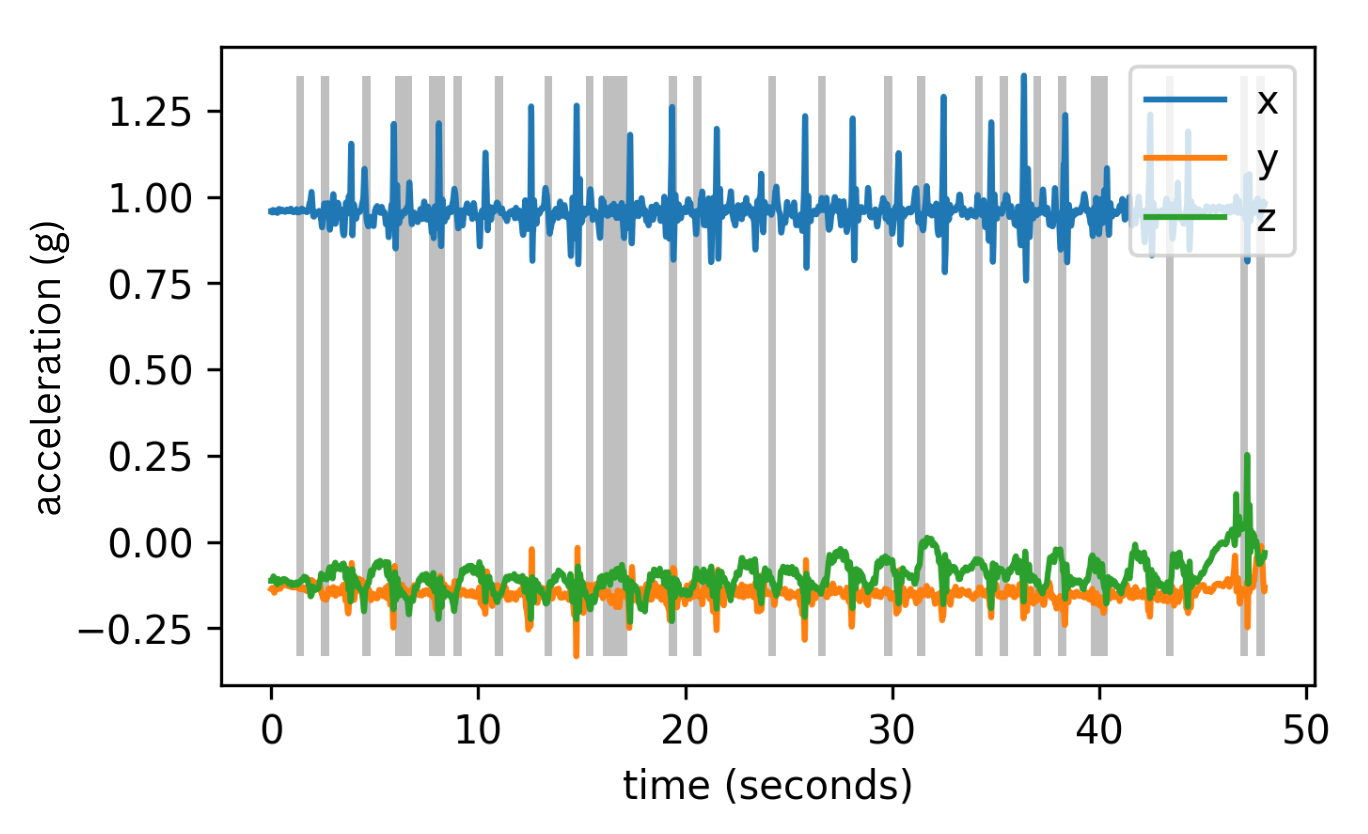}
\caption{The masked operation. This example shows a mask ratio of 0.2 for \textit{heels up/down}. The shaded area was masked.}
\label{fig:mask}
\end{figure}

As outlined in \cite{haresamudram2020masked}, signal reconstruction involved the application of two fully-connected layers. During the training of the unsupervised block, a portion of the input signals was randomly masked. The masking operation was executed as depicted in Fig.~\ref{fig:mask}. As clarified in Section~\ref{subsec:transformer}, the signals were segmented into sections with a length of 800 samples. Each input segment was then further divided into smaller patches with a length of 40 samples. These patches were randomly chosen and masked, resulting in each segment obtaining a boolean vector representing the mask.

When reconstructing the signals, the Mean Squared Error (MSE) function was used as the loss function:

\begin{equation}
L_{MSE}=\frac{1}{NT}\sum_{n,t}m_{t,n}(x_{t,n}-\hat{x}_{t,n})^2,
\end{equation}

where $x_{t,n}$ and $\hat{x}_{t,n}$ denote the true input signals and reconstructed signals at timestamp t and channel n. $T$ and $N$ are the number of samples and channels. Since only the masked samples were considering in the loss function, $m_{t,n}$ denotes index of masked samples. \par

\subsection{Semi-supervised approach}

The supervised and unsupervised blocks were trained simultaneously, and the Transformers shared the weights. Therefore, the loss function was defined as:

\begin{equation}
L = \eta L_{CE}+ L_{MSE}.
\end{equation}

The value of $\eta$ served as a hyperparameter influencing the results. As fine-tuned in the preceding study \cite{hermans2023multi}, it was recommended to set $\eta$ as a large value. However, the primary objective was not to achieve optimized results for the reconstruction task. Reconstruction was only employed to enhance the classification performance. In this study, after conducting multiple experiments, the optimal value for $\eta$ was determined to be 500.

\subsection{Baseline methods}

Several autoencoders were employed as baseline models for comparison with the proposed masked Transformer model. These models were implemented as sequence-to-sequence models, as a part of the architecture in Fig.~\ref{fig:semi structure}. At the output of the latent space, the identical OS-TCN model was connected to generate the classified micro activities.

\paragraph{Transformer SS}
To assess the impact of the masked approach, a semi-supervised Transformer model (Transformer SS) was employed without masked input. The architecture remained consistent with the proposed masked Transformer.

\paragraph{Convolutional Autoencoder (CAE)}
As suggested in a prior study \cite{mohd2021feature}, four convolutional layers (encoder) were utilized to extract latent features, and four deconvolutional layers (decoder) were employed for signal reconstruction. The latent features were subsequently utilized for the classification task after upsampling.

\paragraph{LSTM Autoencoder (LSTMAE)}
The architecture was proposed by \cite{nguyen2021forecasting}. Following the first LSTM layer, information was stored in the cell. Subsequently, another LSTM layer was applied to expand the features into the reconstructed signals. The output features of the LSTM layer could then be utilized for classification.

\paragraph{TCN Autoencoder (TCNAE)}
As proposed by \cite{thill2021temporal}, this was an enhanced CAE architecture. It incorporated stacked dilated convolutional layers as the first TCN block. Subsequently, a max-pooling layer was applied to extract latent features. A subsequent TCN block was then employed for signal reconstruction.

In our last study \cite{shang2024ds}, the OS-TCN outperformed other supervised approaches (as adding more stages did not show improved results). Consequently, it was also employed as a baseline model using the supervised method.

\subsection{Data split}

The Leave-One-Subject-Out Cross-Validation (LOSOCV) method was employed on the data of subjects from the lab. The impact of the mask ratio was explored using the lab-based dataset. Subsequently, the proposed masked Transformer method was compared to the baseline models. Finally, the model was generalized to the home-based dataset using LOSOCV, incorporating the lab-based data into the training set.

\subsection{Evaluation metrix}

\subsubsection{sample-wise evaluation}

The sample-wise evaluation calculated f1-scores based on the results of each sample. For each class c, the f1-score was calculated as:
\begin{equation}
f1_{c}=2*\frac{precision_{c}*recall_{c}}{precision_{c}+recall_{c}},
\label{f:f1}
\end{equation}

where
\begin{equation}
precision_{c}=\frac{TP_{c}}{TP_{c}+FP_{c}}, and
\label{f:precision}
\end{equation}
\begin{equation}
recall_{c}=\frac{TP_{c}}{TP_{c}+FN_{c}}.
\label{f:recell}
\end{equation}

where $TP_{c}$, $FP_{c}$, and $FN_{c}$ denote the numbers of true positive, false positive, and false negative labels from the sample-wise classification output.\par

\subsubsection{segment-wise evaluation}

As detailed in \cite{shang2023otago} and \cite{shang2024ds}, segmental f1-scores of Intersection over Union (IoU) were computed for each class to assess over-segmentation errors. Based on a pre-defined threshold, segment-wise true positive ($TP_{\text{seg}}$), false positive ($FP_{\text{seg}}$), and false negative ($FN_{\text{seg}}$) were defined as follows:

\begin{itemize}
	\item $TPseg$: IoU$\geq$threshold
	\item $FPseg$: IoU$<$threshold, true segments shorter than predicted segments
	\item $FNseg$: IoU$<$threshold, true segments longer than predicted segments
\end{itemize}
	
According to equations~\eqref{f:f1},~\eqref{f:precision},and~\eqref{f:recell}, the segment-wise evaluation matrices could be calculated. In this study, a high threshold of 0.75 was selected, since the micro labels were short and required high classification performance.

\subsubsection{micro activity counts}

From a clinical perspective, counting the number of repetitions for each type of exercise is crucial to monitor the increasing difficulty for the subjects. For each subject, the number of micro activities was tallied for each class. By comparing the predicted and true counts, limits of agreement ($LOA$) for each class $c$ could be calculated using:

\begin{equation}
LOA_{c}= Mean_c \pm 2*Std_c,
\label{f:LOA}
\end{equation}

where $Mean_c$ and $Std_c$ denote the mean and standard deviation of the difference between the true and predicted counts over the subjects.

\subsection{Velocity of chair rising}

Following the identification of micro activities, the acceleration signals of each \textit{chair rising} repetition could be extracted. In accordance with antecedent investigations \cite{marques2020accelerometer}, the velocity corresponding to the \textit{sit-to-stand} and \textit{stand-to-sit} exercises was subsequently derived from the acceleration segments. Subsequent to the application of low-pass filtering with a passband frequency of 20Hz, the gravitational component inherent in the vertical acceleration signals was attenuated:

\begin{equation}
a_x' = a_x - g',
\label{f:acc}
\end{equation}

where $a_x$ and $a_x'$ are the original and modified vertical acceleration. Since the gravitational component could be projected onto different axes of the sensor reference system due to the placement of the sensor and the posture of the subjects, the value of $g'$ was not exactly 9.8$m/s^2$. Therefore, $g'$ was estimated by observing the acceleration when the subject was sitting still.\par

Then, before each sequence of \textit{chair rising} exercise, an initial point had to be manually selected where the acceleration was approximately static. At this point, both acceleration $a_{x,0}$ and velocity $v_{x,0}$ were assumed to be zero. From this initial point, the velocity of the following points could be calculated by:

\begin{equation}
v_{x,t} = \sum_{t}a_{x,t}*dt
\label{f:LOA}
\end{equation}

The time interval $dt$ was selected as 0.01 as the sampling frequency was 100Hz. \par

The computed velocity was anticipated to exhibit a degree of inaccuracy, attributable to factors such as the manually designated initial point and the presence of noise within the data. Nevertheless, the resultant values were deemed sufficient for capturing the overall intensity exhibited by subjects during the \textit{chair rising} activity. This information proves instrumental in facilitating the analysis of their respective fitness levels.

\section{Results}
\label{sec:results}

\subsection{Influence of the mask ratio}

\begin{figure}[!h]
\centering
\includegraphics[width=\columnwidth]{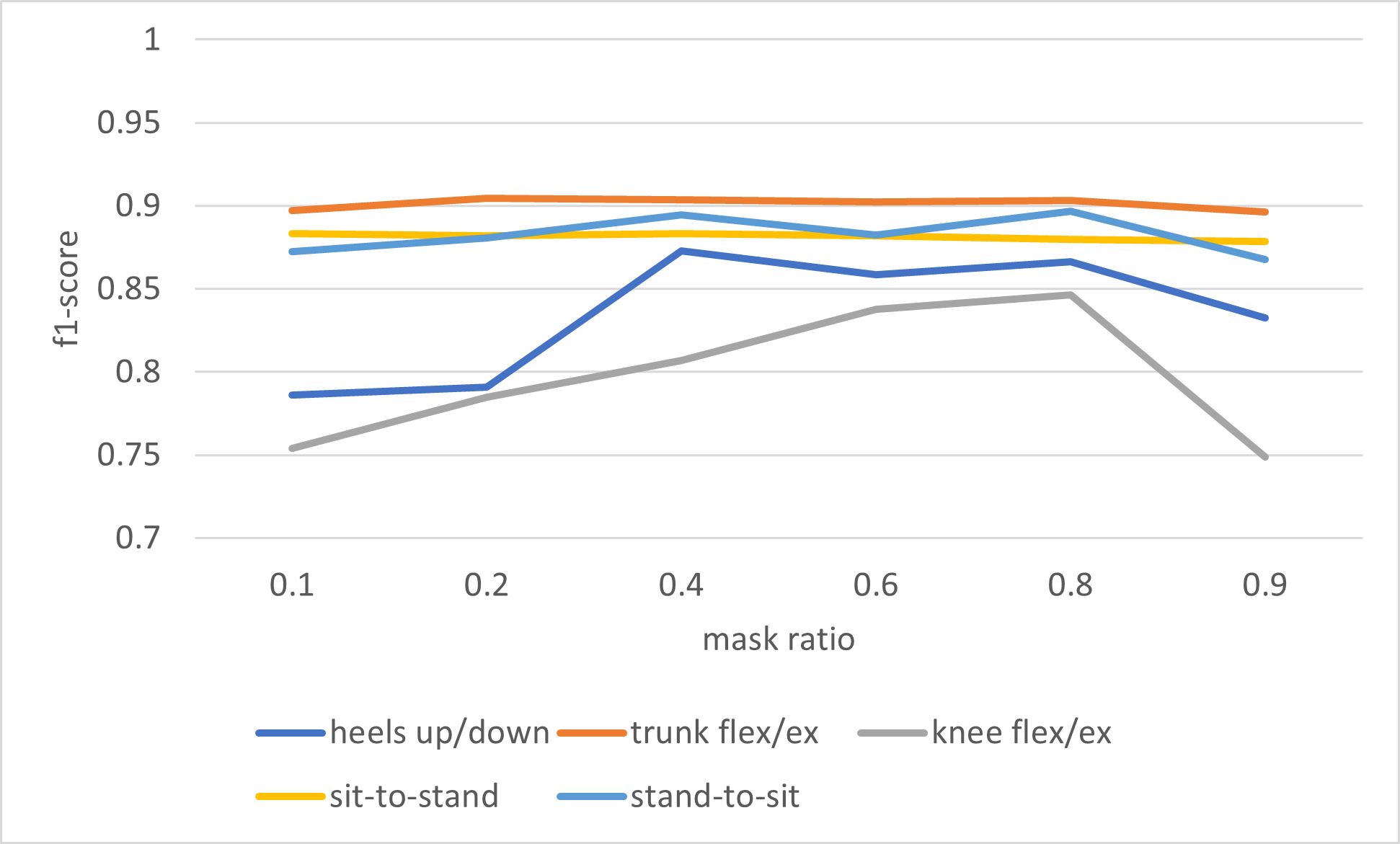}
\caption{The impact of mask ratio on the f1-scores. Abbreviations: flex/ex=flexion/extension}
\label{fig:maskratio}
\end{figure}

Fig.~\ref{fig:maskratio} depicts the f1-scores obtained on the lab-based dataset for the proposed approach, employing various mask ratios in signal reconstruction. Noteworthy observations include the stability of f1-scores for \textit{trunk flexion/extension}, \textit{sit-to-stand} and \textit{stand-to-sit} across varying mask ratios. In contrast, \textit{heels up/down} and \textit{knees flexion/extension} exhibited optimal f1-scores at mask ratios of 0.4 and 0.8, respectively. Beyond these ratios, whether higher than 0.8 or lower than 0.4, a considerable decline in f1-scores was evident. Consequently, for optimal overall f1-scores, a mask ratio of 0.8 was judiciously chosen.

\subsection{Approach comparison}

\begin{figure}[!h]
\centering
\includegraphics[width=\columnwidth]{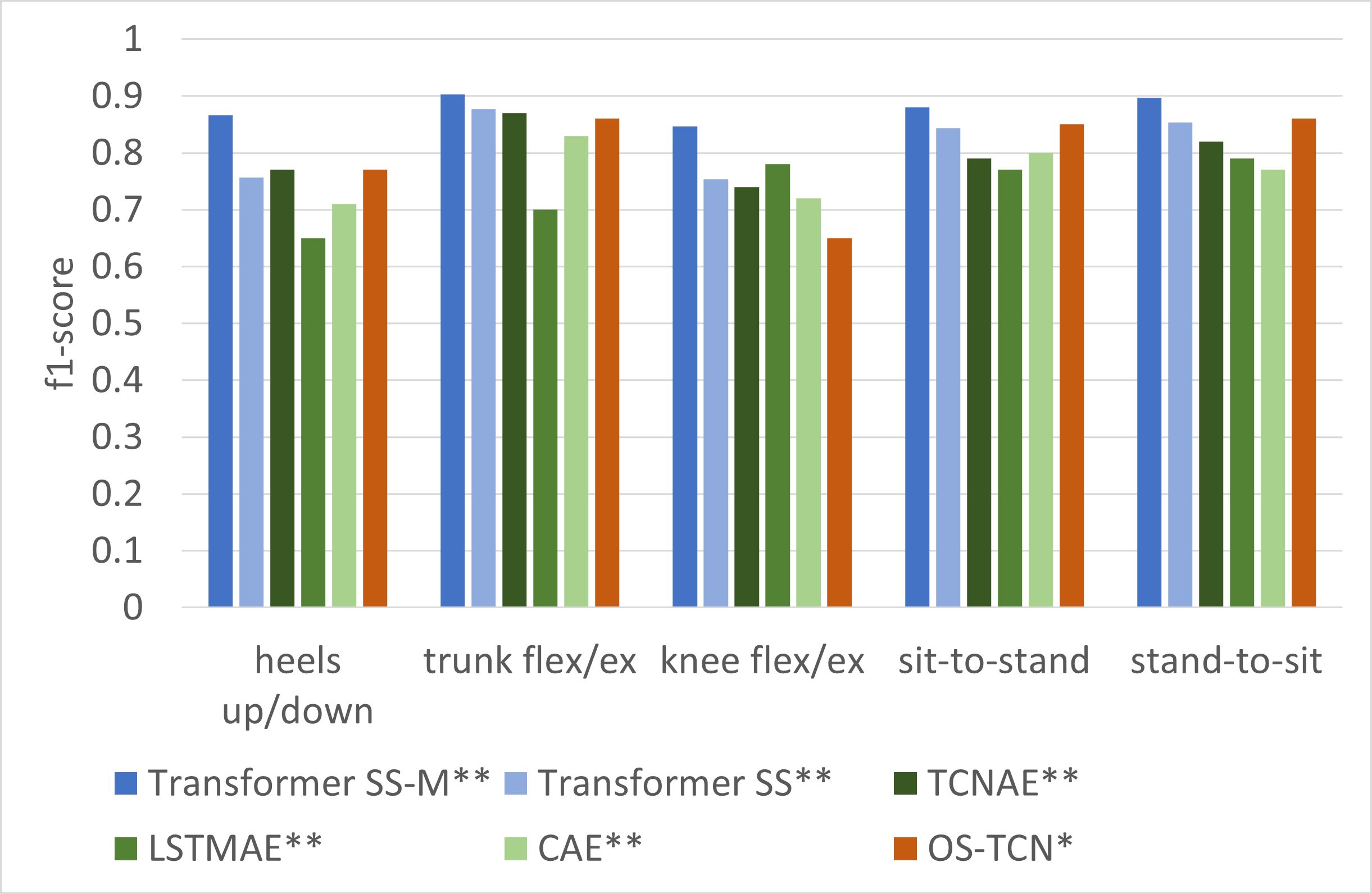}
\caption{The f1-scores of the proposed Transformer SS-M approach and other baseline approaches (*=supervised ** = semi-supervised)}
\label{fig:comparison}
\end{figure}

Fig.~\ref{fig:comparison} shows the f1-scores on the lab-based dataset by different models. Out of the four autoencoder models, the Transformer obtained the highest overall f1-scores. Compared with the supervised method (OS-TCN), however, the Transformer only outperformed on \textit{knees flexion/extension}. After applying the masked approach, the Transformer obtained higher f1-scores than the other baseline methods. Compared with OS-TCN, the f1-scores of \textit{heels up/down} and \textit{knees flexion/extension} obtained the highest increases (0.87 and 0.85, respectively).\par

\subsection{Results in lab vs. home}

\begin{figure}[!h]
\centering
\includegraphics[width=\columnwidth]{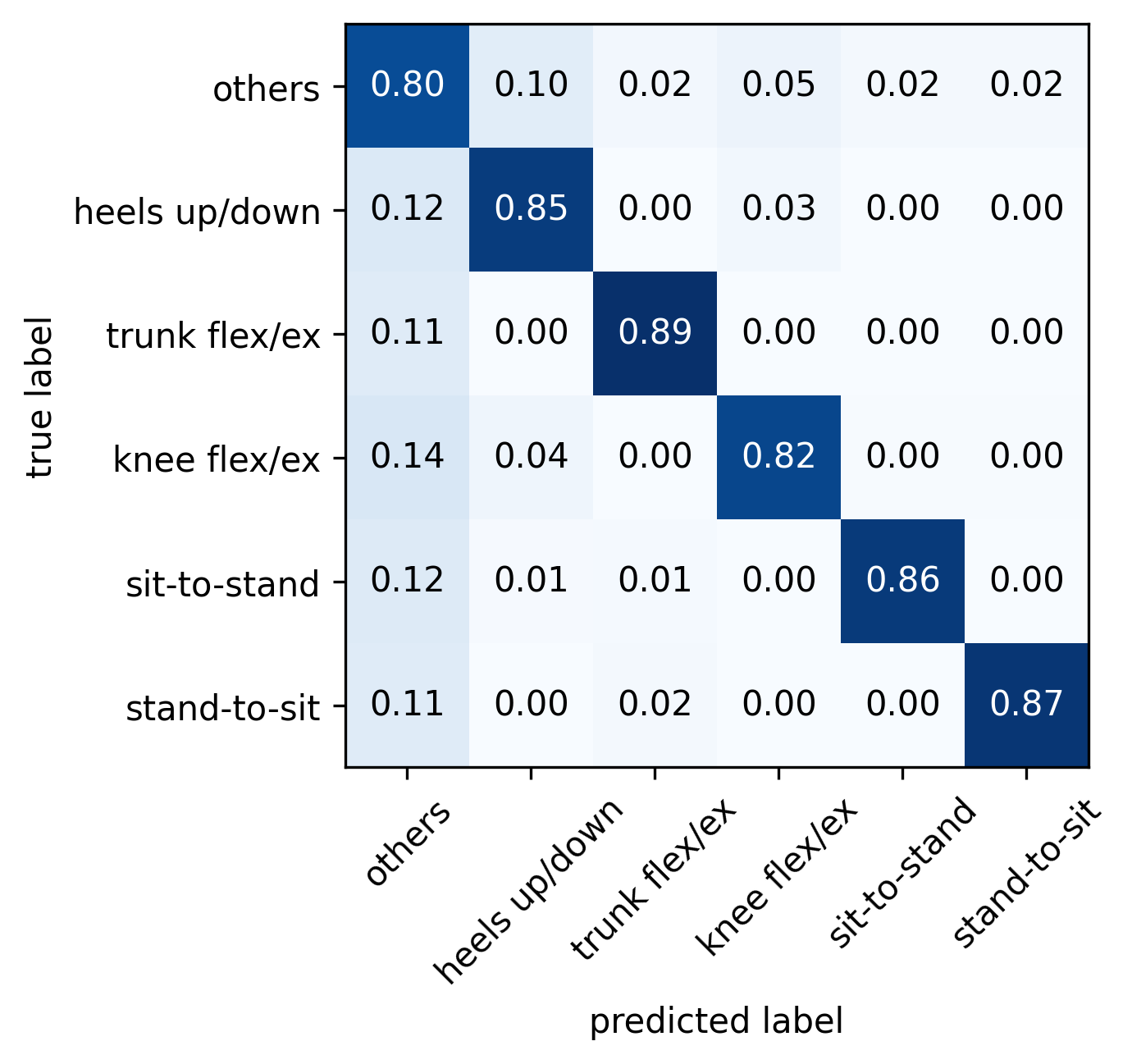}
\caption{The confusion matrix in the lab settings. The values are normalized over the numbers of true labels.}
\label{fig:cm}
\end{figure}

Table~\ref{tab:labhome} provides an overview of the evaluation scores for both the lab-based dataset and the home-based dataset. Notably, in the home settings, the Limits of Agreement (LOA) values were omitted from reporting due to the limited number of subjects (seven). Additionally, only three out of the seven subjects adhered to the \textit{sit-to-stand} and \textit{stand-to-sit} tasks, rendering the results for these two classes less reliable. The confusion matrix in the lab settings is shown in Fig.~\ref{fig:cm}. \par

\begin{table}[!h]
\caption{The evaluation results of the lab-based and home-based dataset}
\label{tab:labhome}
\center
\begin{tabular}{llll}
\hline
\multicolumn{4}{c}{lab}                                                                                                                                                                 \\ \hline
                              & f1-scores                & IoU f1-scores               & \begin{tabular}[c]{@{}l@{}}LOA\\ (number of \\ total true segments)\end{tabular}               \\ \hline
Ankle plan                    & 0.87                     & 0.58                        & -14.98$\pm$22.46 (905)                                                                         \\ \hline
Abdominal                     & 0.90                     & 0.85                        & -0.69$\pm$1.61 (183)                                                                           \\ \hline
Knee bends                    & 0.85                     & 0.75                        & -2.43$\pm$11.67 (692)                                                                          \\ \hline
Sit-to-stand                  & 0.88                     & 0.80                        & -0.79$\pm$2.43 (339)                                                                           \\ \hline
Stand-to-sit                  & 0.90                     & 0.86                        & -0.61$\pm$2.03 (339)                                                                           \\ \hline
\multicolumn{4}{c}{home}                                                                                                                                                                \\ \hline
                              & f1-scores                & IoU f1-scores               & LOA*                                                                                           \\ \hline
Ankle plan                    & 0.81                     & 0.22                        &                                                                                                \\ \hline
Abdominal                     & 0.80                     & 0.55                        &                                                                                                \\ \hline
Knee bends                    & 0.88                     & 0.48                        &                                                                                                \\ \hline
Sit-to-stand**                & 0.38                     & 0.33                        &                                                                                                \\ \hline
Stand-to-sit**                & 0.53                     & 0.28                        &                                                                                                \\ \hline
\multicolumn{4}{l}{\begin{tabular}[c]{@{}l@{}}*The small number of subjects did not support statistical analysis\\ ** Only three subjects performed \textit{chair rising}\end{tabular}}
\end{tabular}
\end{table}

The LOA values exhibited minimal errors and low variance for \textit{trunk flexion/extension}, \textit{sit-to-stand}, and \textit{stand-to-sit}. Across all exercises, a consistent trend was observed wherein the model tended to overestimate the number of repetitions, as evidenced by negative mean values.\par

The f1-scores obtained in both the lab setting and home setting surpassed the clinically applicable threshold of 0.8 \cite{dedeyne_exploring_2021}. Despite a decline in the f1-scores for \textit{heels up/down}, \textit{knees flexion/extension}, and \textit{trunk flexion/extension} when transitioning from the lab to home settings, these scores remained above the clinically relevant threshold of 0.8.\par

\begin{figure}[!h]
\centering
\subfloat[lab]{\includegraphics[width=\columnwidth]{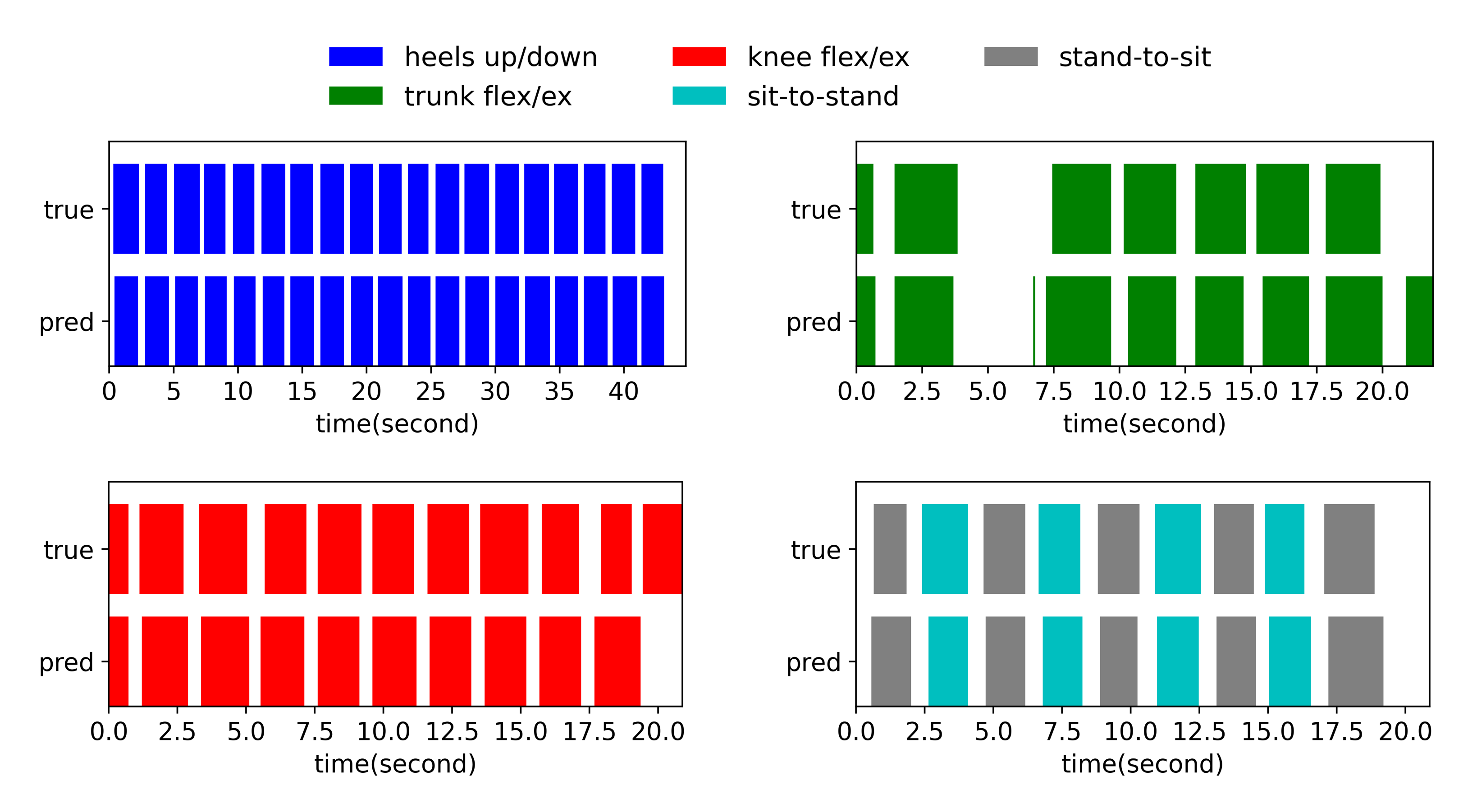}%
}
\hfill
\subfloat[home]{\includegraphics[width=\columnwidth]{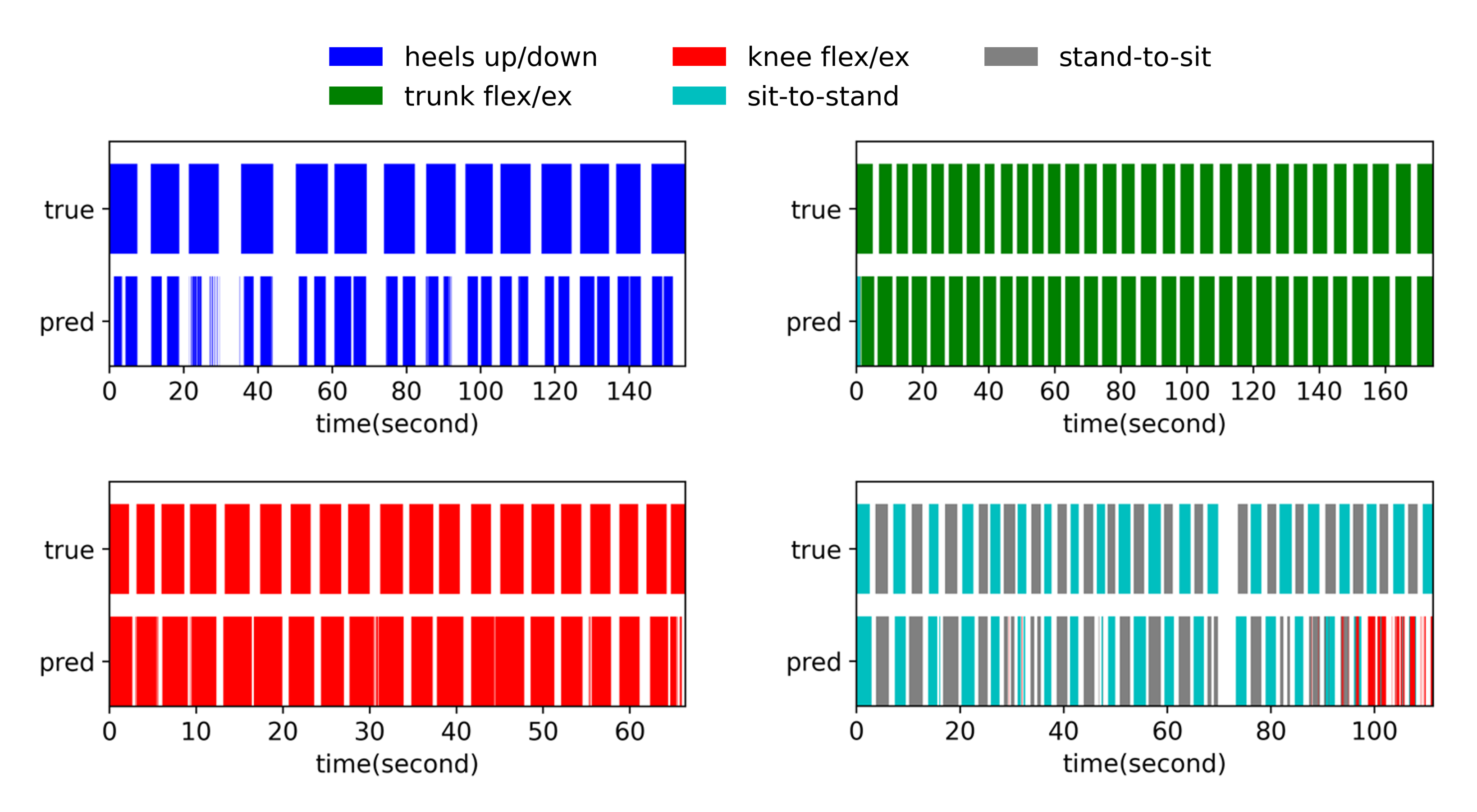}%
}
\caption{The true and predicted labels in lab and home}
\label{fig:labelresults}
\end{figure}

For a visual representation of the predicted micro activities, Fig.~\ref{fig:labelresults} illustrates the true and predicted labels in both lab and home settings. The visualization aligns with the findings that the IoU f1-scores diminished in the home settings. This reduction can be attributed to a single repetition being identified as more shorter segments (more over-segmentation errors). Consequently, the model tended to predict a higher count in the home setting due to this segmentation characteristic. The classification performances for \textit{chair rising} exhibited a decline for the last repetitions.

\subsection{Velocity of chair rising}

\begin{figure}[!h]
\centering
\includegraphics[width=\columnwidth]{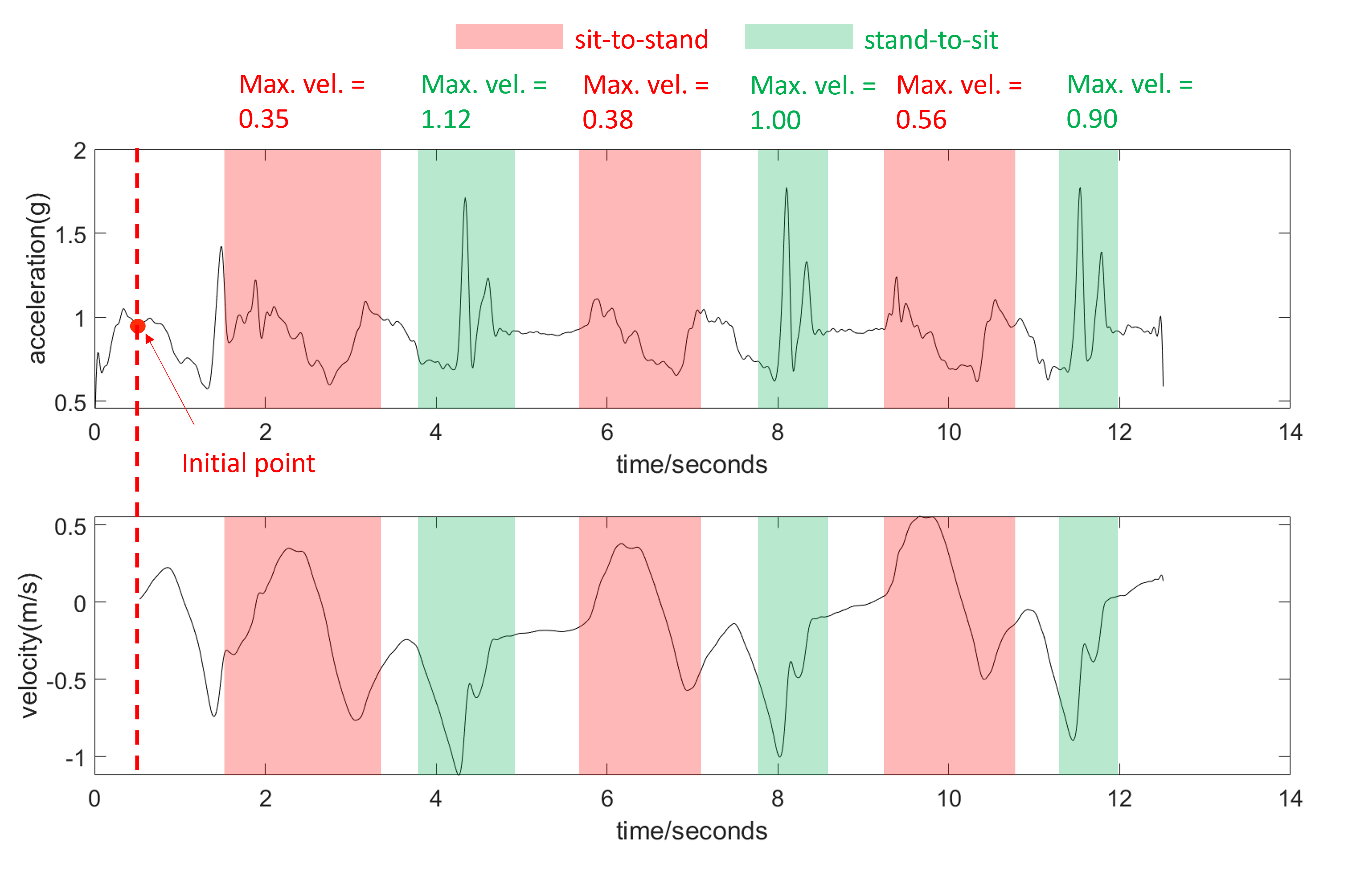}
\caption{The vertical acceleration and velocity of three \textit{chair rising} repetitions. The initial point was selected for integrating the acceleration. The shaded blocks are the predicted labels of \textit{sit-to-stand} and \textit{stand-to-sit} recognized by the proposed approach. The maximum velocity can be calculated for each repetition.}
\label{fig:vel}
\end{figure}

Fig.~\ref{fig:vel} shows an example of the calculated velocity of \textit{sit-to-stand} and \textit{stand-to-sit}. From the raw vertical acceleration signals, the velocity was calculated by integration from a stable initial point. The velocity values could be plotted as a time series. According to the previous steps, the start point and end point of each micro activity could be recognized. By locating the periods of each repetition, the maximum velocity and duration of these exercises could be extracted. Besides, the duration for each repetition could also be extracted.

\section{Discussion}
\label{sec:discussion}

\subsection{The proposed masked Transformer}

The motivation for using semi-supervised learning was the lack of training examples of the micro activities. When annotating each repetition, the workload was substantial since the activities and intervals were very short. Leveraging semi-supervised learning with a focus on multiple tasks allowed the model to glean informative features by reconstructing the signals. However, in this study, the autoencoders did not consistently outperform the supervised learning method. The CAE and LSTMAE yielded lower f1-scores in comparison to the OS-TCN model. These outcomes align with findings in the prior study \cite{haresamudram2020masked}, proving the improvement by adding the masked reconstruction process. \par

The semi-supervised learning approach proposed in this study shared similarities with unsupervised learning methods as documented in the previous work \cite{haresamudram2020masked}. However, a key distinction lies in the methodology. While unsupervised learning methods typically involve a two-step process comprising a reconstruction phase followed by a classification phase, the semi-supervised method presented in this study integrated both supervised and unsupervised loss functions. This simultaneous training allowed for the extraction of latent features based on both processes simultaneously. \par

The incorporation of the masked approach further enhanced the model's generalization ability. By excluding a portion of the training examples, the reconstruction process's complexity increased. While the conventional approach for language models typically applied a low mask ratio \cite{nozza2020mask, wettig2022should}, our study yielded results analogous to image reconstruction studies \cite{he2022masked}, indicating that a higher mask ratio was more effective. This can be attributed to the shared characteristic between IMU signals and images: both possess sparse information. In essence, the intrinsic information embedded in IMU samples tends to be redundant. Consequently, the removal of only a small percentage of samples does not significantly impact the reconstruction process.  \par

\subsection{Lab vs. home settings}

As discussed in our previous study \cite{shang2023otago, shang2024ds}, the evaluation scores in the home setting were decreased compared with the lab setting. This decline was attributed to participants at home lacking instructions or assistance from researchers, leading to a deviation from the prescribed standard movements of the OEP. It was also noted that as the number of repetitions within a session increased, there was a decline in classification performance. This phenomenon could be attributed to the subjects experiencing fatigue, leading to difficulty in adhering to the prescribed exercise standards. Moreover, the exercises conducted at home were often performed at a considerably lower pace by some participants, thereby introducing a heightened intra-subject variance in the signals. However, in this study, the seven recruited subjects at home might not cover such variance. To address this limitation in future studies, a larger pool of subjects engaged in home-based activities should be incorporated during model training, enabling better coverage of variance in the data. \par

In the home setting, the over-segmentation errors were high, especially for \textit{heels up/down}. This limitation was further exacerbated by the choice of sensor placement. Despite user-friendliness, the IMU affixed to the waist, inherently lacked sensitivity to movements specifically originating from the ankles. Recognizing this constraint, future studies should consider incorporating additional sensors, such as placing one directly on the ankle, to capture a more comprehensive set of movement information. 

\subsection{Clinical relevance}

The initial clinical relevance of the study involved quantifying the number of certain OEP sub-classes. Typically, older adults practicing OEP more frequently can achieve greater repetitions for each exercise. In this study, the assessment of counts was stricter than that of f1-scores and IoU f1-scores. During counting, each over-segmentation error was considered a repetition, contributing to higher variance for some exercises. This issue could potentially be addressed by implementing simple algorithms, such as discarding extremely short segments. Moreover, for exercises like \textit{sit-to-stand} and \textit{stand-to-sit}, the repetitions could be carried out in daily life, although in the absence of ground truth labels. In the future, it is anticipated that the system will undergo validation not only within the context of OEP but also beyond the program's scope.
 \par

The second clinical relevance pertained to the computation of the velocity during \textit{sit-to-stand} and \textit{stand-to-sit}. Previous studies had also calculated velocity based on individually recorded and processed repetitions, necessitating manual annotation of start and end points. However, this manual approach was deemed unrealistic for daily life applications. In contrast, our study employed a system capable of recognizing the start and end points for each repetition with high f1-scores. This made it feasible to automatically monitor velocity throughout the day, eliminating the need for manual intervention in annotating the start and end points.

\section{Conclusion}
\label{sec:conclusion}

The study demonstrates the viability of recognizing individual short repetitions within OEP, a rehabilitation program designed for older adults. The clinical outcomes derived from this recognition have practical applications for therapists. The introduced masked semi-supervised approach provides a potential solution for HAR systems grappling with limited dataset sizes. Moving forward, there is potential for the system to be applied in broader clinical contexts and other HAR applications.\par

Additionally, enhancing the recognition of OEP could be achieved by incorporating more sensors to monitor more exercises. This expansion in sensor coverage holds promise for further improving the accuracy and effectiveness of the recognition process.

\section*{Acknowledgment}

This work was supported by the China Scholarship Council (CSC).\par

The ENHANce project (S60763) received a junior research project grant from the Research Foundation Flanders (FWO) (G099721N). The funding provider did not contribute or influence the design of the study and data collection, analysis and interpretation in writing this manuscript.

\section*{REFERENCES}

\bibliographystyle{ieeetr}
\bibliography{generic-color}

\begin{thebibliography}{10}

\bibitem{thomas_does_2010}
S.~Thomas, S.~Mackintosh, and J.~Halbert, ``Does the ‘{Otago} exercise
  programme’ reduce mortality and falls in older adults?: a systematic review
  and meta-analysis,'' {\em Age and Ageing}, vol.~39, pp.~681--687, Nov. 2010.

\bibitem{mat_effect_2018}
S.~Mat, C.~T. Ng, P.~J. Tan, N.~Ramli, F.~Fadzli, F.~I. Rozalli, M.~Mazlan,
  K.~D. Hill, and M.~P. Tan, ``Effect of {Modified} {Otago} {Exercises} on
  {Postural} {Balance}, {Fear} of {Falling}, and {Fall} {Risk} in {Older}
  {Fallers} {With} {Knee} {Osteoarthritis} and {Impaired} {Gait} and {Balance}:
  {A} {Secondary} {Analysis},'' {\em PM\&R}, vol.~10, no.~3, pp.~254--262,
  2018.

\bibitem{almarzouki_improved_2020}
R.~Almarzouki, G.~Bains, E.~Lohman, B.~Bradley, T.~Nelson, S.~Alqabbani,
  A.~Alonazi, and N.~Daher, ``Improved balance in middle-aged adults after 8
  weeks of a modified version of otago exercise program: A randomized
  controlled trial,'' {\em Plos one}, vol.~15, no.~7, p.~e0235734, 2020.

\bibitem{shang2023otago}
M.~Shang, L.~Dedeyne, J.~Dupont, L.~Vercauteren, N.~Amini, L.~Lapauw,
  E.~Gielen, S.~Verschueren, C.~Varon, W.~De~Raedt, and B.~Vanrumste, ``Otago
  exercises monitoring for older adults by a single imu and hierarchical
  machine learning models,'' {\em IEEE Transactions on Neural Systems and
  Rehabilitation Engineering}, vol.~32, pp.~462--471, 2024.

\bibitem{dedeyne_exploring_2021}
L.~Dedeyne, J.~A. Wullems, J.~Dupont, J.~Tournoy, E.~Gielen, and
  S.~Verschueren, ``Exploring machine learning models based on accelerometer
  sensor alone or combined with gyroscope to classify home-based exercises and
  physical behavior in (pre) sarcopenic older adults,'' {\em Journal for the
  Measurement of Physical Behaviour}, vol.~4, no.~2, pp.~174--186, 2021.

\bibitem{bevilacqua_human_2019}
A.~Bevilacqua, K.~MacDonald, A.~Rangarej, V.~Widjaya, B.~Caulfield, and
  T.~Kechadi, ``Human activity recognition with convolutional neural
  networks,'' in {\em Joint European Conference on Machine Learning and
  Knowledge Discovery in Databases}, pp.~541--552, Springer, 2019.

\bibitem{shang2024ds}
M.~Shang, L.~Dedeyne, J.~Dupont, L.~Vercauteren, N.~Amini, L.~Lapauw,
  E.~Gielen, S.~Verschueren, C.~Varon, W.~De~Raedt, {\em et~al.}, ``Ds-ms-tcn:
  Otago exercises recognition with a dual-scale multi-stage temporal
  convolutional network,'' {\em arXiv preprint arXiv:2402.02910}, 2024.

\bibitem{kocic2018effectiveness}
M.~Kocic, Z.~Stojanovic, D.~Nikolic, M.~Lazovic, R.~Grbic, L.~Dimitrijevic, and
  M.~Milenkovic, ``The effectiveness of group otago exercise program on
  physical function in nursing home residents older than 65 years: A randomized
  controlled trial,'' {\em Archives of gerontology and geriatrics}, vol.~75,
  pp.~112--118, 2018.

\bibitem{martinez2019probabilistic}
U.~Martinez-Hernandez and A.~A. Dehghani-Sanij, ``Probabilistic identification
  of sit-to-stand and stand-to-sit with a wearable sensor,'' {\em Pattern
  Recognition Letters}, vol.~118, pp.~32--41, 2019.

\bibitem{adamowicz2020assessment}
L.~Adamowicz, F.~I. Karahanoglu, C.~Cicalo, H.~Zhang, C.~Demanuele,
  M.~Santamaria, X.~Cai, and S.~Patel, ``Assessment of sit-to-stand transfers
  during daily life using an accelerometer on the lower back,'' {\em Sensors},
  vol.~20, no.~22, p.~6618, 2020.

\bibitem{marques2020accelerometer}
D.~L. Marques, H.~P. Neiva, I.~M. Pires, D.~A. Marinho, and M.~C. Marques,
  ``Accelerometer data from the performance of sit-to-stand test by elderly
  people,'' {\em Data in brief}, vol.~33, p.~106328, 2020.

\bibitem{soekhoe2016impact}
D.~Soekhoe, P.~Van Der~Putten, and A.~Plaat, ``On the impact of data set size
  in transfer learning using deep neural networks,'' in {\em Advances in
  Intelligent Data Analysis XV: 15th International Symposium, IDA 2016,
  Stockholm, Sweden, October 13-15, 2016, Proceedings 15}, pp.~50--60,
  Springer, 2016.

\bibitem{hermans2023multi}
T.~Hermans, L.~Smets, K.~Lemmens, A.~Dereymaeker, K.~Jansen, G.~Naulaers,
  F.~Zappasodi, S.~Van~Huffel, S.~Comani, and M.~De~Vos, ``A multi-task and
  multi-channel convolutional neural network for semi-supervised neonatal
  artefact detection,'' {\em Journal of Neural Engineering}, vol.~20, no.~2,
  p.~026013, 2023.

\bibitem{wang_survey_2019}
Y.~Wang, S.~Cang, and H.~Yu, ``A survey on wearable sensor modality centred
  human activity recognition in health care,'' {\em Expert Systems with
  Applications}, vol.~137, pp.~167--190, Dec. 2019.

\bibitem{8684824}
J.~Huang, S.~Lin, N.~Wang, G.~Dai, Y.~Xie, and J.~Zhou, ``Tse-cnn: A two-stage
  end-to-end cnn for human activity recognition,'' {\em IEEE Journal of
  Biomedical and Health Informatics}, vol.~24, no.~1, pp.~292--299, 2020.

\bibitem{lee_human_2017}
S.-M. Lee, S.~M. Yoon, and H.~Cho, ``Human activity recognition from
  accelerometer data using {Convolutional} {Neural} {Network},'' in {\em 2017
  {IEEE} {International} {Conference} on {Big} {Data} and {Smart} {Computing}
  ({BigComp})}, pp.~131--134, Feb. 2017.
\newblock ISSN: 2375-9356.

\bibitem{wagner_activity_2017}
D.~Wagner, K.~Kalischewski, J.~Velten, and A.~Kummert, ``Activity recognition
  using inertial sensors and a 2-{D} convolutional neural network,'' in {\em
  2017 10th {International} {Workshop} on {Multidimensional} ({nD}) {Systems}
  ({nDS})}, pp.~1--6, Sept. 2017.

\bibitem{zhao_deep_2018}
Y.~Zhao, R.~Yang, G.~Chevalier, X.~Xu, and Z.~Zhang, ``Deep {Residual}
  {Bidir}-{LSTM} for {Human} {Activity} {Recognition} {Using} {Wearable}
  {Sensors},'' {\em Mathematical Problems in Engineering}, vol.~2018,
  p.~e7316954, Dec. 2018.
\newblock Publisher: Hindawi.

\bibitem{mutegeki_cnn-lstm_2020}
R.~Mutegeki and D.~S. Han, ``A {CNN}-{LSTM} {Approach} to {Human} {Activity}
  {Recognition},'' in {\em 2020 {International} {Conference} on {Artificial}
  {Intelligence} in {Information} and {Communication} ({ICAIIC})},
  pp.~362--366, Feb. 2020.

\bibitem{mekruksavanich_smartwatch-based_2020}
S.~Mekruksavanich and A.~Jitpattanakul, ``Smartwatch-based {Human} {Activity}
  {Recognition} {Using} {Hybrid} {LSTM} {Network},'' in {\em 2020 {IEEE}
  {SENSORS}}, pp.~1--4, Oct. 2020.
\newblock ISSN: 2168-9229.

\bibitem{farha_ms-tcn_2019}
Y.~A. Farha and J.~Gall, ``Ms-tcn: Multi-stage temporal convolutional network
  for action segmentation,'' in {\em Proceedings of the IEEE/CVF Conference on
  Computer Vision and Pattern Recognition (CVPR)}, June 2019.

\bibitem{vaswani2017attention}
A.~Vaswani, N.~Shazeer, N.~Parmar, J.~Uszkoreit, L.~Jones, A.~N. Gomez,
  {\L}.~Kaiser, and I.~Polosukhin, ``Attention is all you need,'' {\em Advances
  in neural information processing systems}, vol.~30, 2017.

\bibitem{vig2019analyzing}
J.~Vig and Y.~Belinkov, ``Analyzing the structure of attention in a transformer
  language model,'' {\em arXiv preprint arXiv:1906.04284}, 2019.

\bibitem{wang2019language}
C.~Wang, M.~Li, and A.~J. Smola, ``Language models with transformers,'' {\em
  arXiv preprint arXiv:1904.09408}, 2019.

\bibitem{chen2021pre}
H.~Chen, Y.~Wang, T.~Guo, C.~Xu, Y.~Deng, Z.~Liu, S.~Ma, C.~Xu, C.~Xu, and
  W.~Gao, ``Pre-trained image processing transformer,'' in {\em Proceedings of
  the IEEE/CVF conference on computer vision and pattern recognition},
  pp.~12299--12310, 2021.

\bibitem{dong2018speech}
L.~Dong, S.~Xu, and B.~Xu, ``Speech-transformer: a no-recurrence
  sequence-to-sequence model for speech recognition,'' in {\em 2018 IEEE
  international conference on acoustics, speech and signal processing
  (ICASSP)}, pp.~5884--5888, IEEE, 2018.

\bibitem{10409509}
M.~Chen, Y.~Li, L.~Zhang, L.~Liu, B.~Han, W.~Shi, and S.~Wei, ``Elimination of
  random mixed noise in ecg using convolutional denoising autoencoder with
  transformer encoder,'' {\em IEEE Journal of Biomedical and Health
  Informatics}, pp.~1--12, 2024.

\bibitem{dirgova2022wearable}
I.~Dirgov{\'a}~Lupt{\'a}kov{\'a}, M.~Kubov{\v{c}}{\'\i}k, and
  J.~Posp{\'\i}chal, ``Wearable sensor-based human activity recognition with
  transformer model,'' {\em Sensors}, vol.~22, no.~5, p.~1911, 2022.

\bibitem{mohd2021feature}
M.~H. Mohd~Noor, ``Feature learning using convolutional denoising autoencoder
  for activity recognition,'' {\em Neural Computing and Applications}, vol.~33,
  no.~17, pp.~10909--10922, 2021.

\bibitem{thill2021temporal}
M.~Thill, W.~Konen, H.~Wang, and T.~B{\"a}ck, ``Temporal convolutional
  autoencoder for unsupervised anomaly detection in time series,'' {\em Applied
  Soft Computing}, vol.~112, p.~107751, 2021.

\bibitem{seyfiouglu2018deep}
M.~S. Seyfio{\u{g}}lu, A.~M. {\"O}zbayo{\u{g}}lu, and S.~Z. G{\"u}rb{\"u}z,
  ``Deep convolutional autoencoder for radar-based classification of similar
  aided and unaided human activities,'' {\em IEEE Transactions on Aerospace and
  Electronic Systems}, vol.~54, no.~4, pp.~1709--1723, 2018.

\bibitem{nguyen2021forecasting}
H.~D. Nguyen, K.~P. Tran, S.~Thomassey, and M.~Hamad, ``Forecasting and anomaly
  detection approaches using lstm and lstm autoencoder techniques with the
  applications in supply chain management,'' {\em International Journal of
  Information Management}, vol.~57, p.~102282, 2021.

\bibitem{8422895}
H.~Zou, Y.~Zhou, J.~Yang, H.~Jiang, L.~Xie, and C.~J. Spanos, ``Deepsense:
  Device-free human activity recognition via autoencoder long-term recurrent
  convolutional network,'' in {\em 2018 IEEE International Conference on
  Communications (ICC)}, pp.~1--6, 2018.

\bibitem{yu2023semi}
H.~Yu, K.~Zhao, and X.~Xu, ``Semi-mae: Masked autoencoders for semi-supervised
  vision transformers,'' {\em arXiv preprint arXiv:2301.01431}, 2023.

\bibitem{nozza2020mask}
D.~Nozza, F.~Bianchi, and D.~Hovy, ``What the [mask]? making sense of
  language-specific bert models,'' {\em arXiv preprint arXiv:2003.02912}, 2020.

\bibitem{wettig2022should}
A.~Wettig, T.~Gao, Z.~Zhong, and D.~Chen, ``Should you mask 15\% in masked
  language modeling?,'' {\em arXiv preprint arXiv:2202.08005}, 2022.

\bibitem{he2022masked}
K.~He, X.~Chen, S.~Xie, Y.~Li, P.~Doll{\'a}r, and R.~Girshick, ``Masked
  autoencoders are scalable vision learners,'' in {\em Proceedings of the
  IEEE/CVF conference on computer vision and pattern recognition},
  pp.~16000--16009, 2022.

\bibitem{chien2022maeeg}
H.-Y.~S. Chien, H.~Goh, C.~M. Sandino, and J.~Y. Cheng, ``Maeeg: Masked
  auto-encoder for eeg representation learning,'' {\em arXiv preprint
  arXiv:2211.02625}, 2022.

\bibitem{haresamudram2020masked}
H.~Haresamudram, A.~Beedu, V.~Agrawal, P.~L. Grady, I.~Essa, J.~Hoffman, and
  T.~Pl{\"o}tz, ``Masked reconstruction based self-supervision for human
  activity recognition,'' in {\em Proceedings of the 2020 ACM International
  Symposium on Wearable Computers}, pp.~45--49, 2020.

\bibitem{cruz2010sarcopenia}
A.~J. Cruz-Jentoft, J.~P. Baeyens, J.~M. Bauer, Y.~Boirie, T.~Cederholm,
  F.~Landi, F.~C. Martin, J.-P. Michel, Y.~Rolland, S.~M. Schneider, {\em
  et~al.}, ``Sarcopenia: European consensus on definition and diagnosisreport
  of the european working group on sarcopenia in older peoplea. j. cruz-gentoft
  et al.,'' {\em Age and ageing}, vol.~39, no.~4, pp.~412--423, 2010.

\bibitem{ibanez2022masked}
D.~Ibanez, R.~Fernandez-Beltran, F.~Pla, and N.~Yokoya, ``Masked auto-encoding
  spectral--spatial transformer for hyperspectral image classification,'' {\em
  IEEE Transactions on Geoscience and Remote Sensing}, vol.~60, pp.~1--14,
  2022.

\bibitem{zerveas2021transformer}
G.~Zerveas, S.~Jayaraman, D.~Patel, A.~Bhamidipaty, and C.~Eickhoff, ``A
  transformer-based framework for multivariate time series representation
  learning,'' in {\em Proceedings of the 27th ACM SIGKDD conference on
  knowledge discovery \& data mining}, pp.~2114--2124, 2021.

\end{thebibliography}

\end{document}